\documentclass[letterpaper]{article} 
\usepackage{aaai25}  
\usepackage{times}  
\usepackage{helvet}  
\usepackage{courier}  
\usepackage[hyphens]{url}  
\usepackage{graphicx} 
\usepackage{natbib}  
\usepackage{caption} 
\usepackage{algorithm}
\usepackage{algorithmic}

\usepackage{newfloat}
\usepackage{listings}

\usepackage{pifont}
\usepackage{subfigure}
\usepackage{booktabs}

\usepackage{amsmath}
\usepackage{amssymb}
\usepackage{mathtools}
\usepackage{amsthm}
\usepackage{multirow}
\usepackage{xcolor}
\definecolor{mycolor}{RGB}{202,222,252}

\newtheorem{theorem}{Theorem}
\newtheorem{lemma}{Lemma}
\newtheorem{assumption}{Assumption}

\DeclareCaptionStyle{ruled}{labelfont=normalfont,labelsep=colon,strut=off} 
\floatstyle{ruled}
\newfloat{listing}{tb}{lst}{}
\floatname{listing}{Listing}
\title{GAS: Generative Activation-Aided Asynchronous Split Federated Learning}
\author{
    Jiarong Yang,
    Yuan Liu
}
\affiliations{
    School of Electronic and Information Engineering, South China University of Technology

    eejryang@mail.scut.edu.cn, eeyliu@scut.edu.cn
}

\usepackage{bibentry}

\begin{document}

\maketitle

\begin{abstract}
Split Federated Learning (SFL) splits and collaboratively trains a shared model between clients and server, where clients transmit activations and client-side models to server for updates. Recent SFL studies assume synchronous transmission of activations and client-side models from clients to server. However, due to significant variations in computational and communication capabilities among clients, activations and client-side models arrive at server asynchronously. The delay caused by asynchrony significantly degrades the performance of SFL.
To address this issue, we consider an asynchronous SFL framework, where an activation buffer and a model buffer are embedded on the server to manage the asynchronously transmitted activations and client-side models, respectively. Furthermore, as asynchronous activation transmissions cause the buffer to frequently receive activations from resource-rich clients, leading to biased updates of the server-side model, we propose Generative activations-aided Asynchronous SFL (GAS). In GAS, the server maintains an activation distribution for each label based on received activations and generates activations from these distributions according to the degree of bias. These generative activations are then used to assist in updating the server-side model, ensuring more accurate updates. We derive a tighter convergence bound, and our experiments demonstrate the effectiveness of the proposed method. The code is available at https://github.com/eejiarong/GAS.
\end{abstract}


%

\section{Introduction}
Split Federated Learning (SFL) \cite{jeon2020privacy,thapa2022splitfed} emerges as a promising solution for efficient resource-constrained distributed learning by combining the benefits of both Federated Learning (FL) \cite{mcmahan2017communication,singh2022federated} and Split Learning (SL) \cite{gupta2018distributed}. Specifically, in SFL, the model is split into two parts: the initial layers are processed in parallel by the participating clients, and the intermediate activations are sent to the server, which completes the remaining layers. The server then sends the backpropagated gradients back to the clients, who use these gradients to update their client-side models. After several iterations, the server aggregates the client-side models to form the globally updated model.
\par
Traditional SFL always assumes synchronous model exchange, where the server waits to receive all client-side models for aggregation. However, since clients have different communication and computational capabilities, client-side models are uploaded at the server asynchronously. The slow clients are referred to as stragglers, delaying the overall training process.
Previous works in FL such as FedAsync \cite{xie2019asynchronous} and FedBuff \cite{nguyen2022federated} are proposed to tackle the stragglers issue by allowing clients to update the global model asynchronously. Additionally, CA$^2$FL \cite{wang2023tackling} addresses convergence degradation by caching and reusing previous updates for global calibration, ensuring more consistent model updates despite asynchronous conditions. 
\par
However, the existing works on stragglers issue in SFL still have limitations. On one hand, the adaptive model splitting methods \cite{yan2023have,shen2023ringsfl} for addressing this issue in SFL are constrained by the model structure. Specifically, these methods attempt to balance the arrival time of activations by selecting appropriate split layers of the model.
Nevertheless, when the sizes of activations output by different model layers are similar or the computational and communication capabilities of clients are highly different, it is impossible to ensure simultaneous arrival of activations, regardless of the chosen split layers.
On the other hand, the stragglers issue is more serious in SFL. Specifically, recent SFL methods \cite{huang2023minibatch,yang2024scala} assume synchronous activation transmissions with heterogeneous client data, where the uploaded activations are concatenated to update the server-side model centrally, reducing the bias in the deep layers of the model \cite{luo2021no}. 
However, these methods require the server to wait for stragglers to send their activations at the end of each local iteration, and the frequent transmissions of activations exacerbate the stragglers issue. 
\par
To address the above issues, we propose the asynchronous learning framework for SFL, where an activation buffer and a model buffer are embedded on the server to handle asynchronous updates. Specifically, the activation buffer stores the activations uploaded asynchronously. When the buffer is full, the server concatenates these activations and uses them to update the server-side model. Similarly, the model buffer stores the client-side models uploaded asynchronously. When the buffer is full, the server aggregates the stored client-side models. By introducing the two buffers, we ensure efficient model updates and reduce delays caused by stragglers. However, due to the heterogeneous communication and computational capabilities of clients, the activation buffer may frequently receive activations from resource-rich clients, leading to biased updates in the server-side model \cite{leconte2024queuing,liu2024fedasmu}. To solve this issue, we propose Generative activation-aided Asynchronous SFL (GAS). Specifically, the server maintains a distribution of activations for each label, dynamically updated based on the uploaded activations. When updating the server-side model, we generate the activations from the distributions according to the degree of bias. Then these generative activations are concatenated with the stored activations to update the server-side model, thereby mitigating the model update bias introduced by stragglers.
We summarize our contributions in this paper as follows:
\begin{itemize}
\item We propose an asynchronous SFL framework that enables the asynchronous transmissions of activations and client-side models. To our best knowledge, this is first attempt considering asynchronous SFL.
\item We propose GAS (Generative activation-aided Asynchronous SFL), where the server updates the activation distribution for each label based on the uploaded activations and generates activations from the distributions to assist server-side model updates, mitigating the model update bias caused by stragglers.
\item Several useful insights are obtained via our theoretical analysis: First, GAS can mitigate the gradient dissimilarity introduced by stragglers. Second, GAS achieves a tighter convergence bound. Third, by setting a decaying learning rate, the impact of stragglers can be gradually mitigated as the training progresses.
\end{itemize}

\section{Related Works}
\subsection{Split Federated Learning}
SFL \cite{thapa2022splitfed} combines the strengths of FL \cite{mcmahan2017communication} and SL \cite{gupta2018distributed} to offer a more efficient and scalable learning framework. Recent research has explored various aspects of SFL. To enhance communication efficiency, FedLite \cite{wang2022fedlite} employ compression techniques to reduce the volume of activation data transmitted. Simultaneously, the work by \cite{han2021accelerating} introduces auxiliary networks on the client side, eliminating the need for sending backpropagated gradients. In terms of privacy preservation, ResSFL \cite{li2022ressfl} and NoPeek \cite{li2022ressfl} implement attacker-aware training with an inversion score regularization term to counteract model inversion attacks. Additionally, the works by \cite{xiao2021mixing} and \cite{thapa2022splitfed} leverage mixed activations and differential privacy to safeguard against privacy breaches from intermediate activations. To optimize performance for heterogeneous clients, SCALA \cite{yang2024scala} and MiniBatch-SFL \cite{huang2023minibatch} employ activation concatenation and implement centralized training on the server, thereby enhancing model robustness and accuracy. Meanwhile, S$^2$FL \cite{yan2023have} and RingSFL \cite{shen2023ringsfl} address the stragglers issue by employing adaptive model splitting methods. Furthermore, recent works \cite{lin2024adaptsfl,xu2023accelerating} refines SFL for real-world communication environments by selecting model split layers based on client channel conditions.

\subsection{Asynchronous Federated Learning}
Asynchronous FL addresses the limitations of traditional synchronous FL in heterogeneous environments, where ``stragglers", or slow clients, can degrade overall training performance and efficiency \cite{wang2021field}. Early asynchronous FL frameworks \cite{xie2019asynchronous,chen2019communication} mitigate the impact of stragglers by adaptively weighting the local updates. ASO-Fed \cite{chen2020asynchronous} employs a dynamic learning strategy to adjust the local training step size, reducing the staleness effects caused by stragglers. FedBuff \cite{nguyen2022federated} introduces a buffering mechanism to temporarily store updates from faster clients, achieving higher concurrency and improving training efficiency. CA$^2$FL \cite{wang2023tackling} further advances this approach by caching and calibrating updates based on data properties to handle both stragglers and data heterogeneity. FedCompass \cite{li2023fedcompass} enhances efficiency by using a computing power-aware scheduler to prioritize updates from more powerful clients, thus reducing the waiting time for stragglers. FedASMU \cite{liu2024fedasmu} addresses the stragglers issue through dynamic model aggregation and adaptive local model adjustment methods. Moreover, some works \cite{lee2021adaptive,wang2022asynchronous,zhu2022online,hu2023scheduling} have further optimizes the performance of asynchronous FL in wireless communication environments through staleness-aware model aggregation and client selection schemes.
\par
Note that the asynchronously transmitted activations are concatenated rather than aggregated to update the server-side model in SFL, introducing unique challenges that make previous asynchronous FL methods inapplicable. Furthermore, the more frequent transmissions of activations exacerbate the stragglers issue. Existing SFL methods \cite{yan2023have,shen2023ringsfl} employ adaptive model splitting to balance activation arrival times; however, they are constrained by the model structure.
The above challenges highlight the need for further research to develop a tailored asynchronous framework for SFL. In this paper, we propose GAS to fill the gap by introducing a novel buffer mechanism and generative activations, which address the stragglers issue in SFL and achieve better model performance.
Additionally, while CCVR \cite{luo2021no} and FedImpro \cite{tang2024fedimpro} also employ activation generation to enhance model performance, they require the additional transmission of local activation distributions. GAS distinguishes itself by leveraging the inherent characteristics of the SFL framework to dynamically update activation distributions using the activations continuously uploaded by clients, without incurring extra communication overhead.

\section{Proposed Method}
In this section, we systematically introduce GAS, which employs an activation buffer and a model buffer to enable asynchronous transmissions of activations and client-side models, while leveraging generative activations to mitigate update bias caused by stragglers.

\begin{figure*}[t]
\centering
\includegraphics[width=1\textwidth]{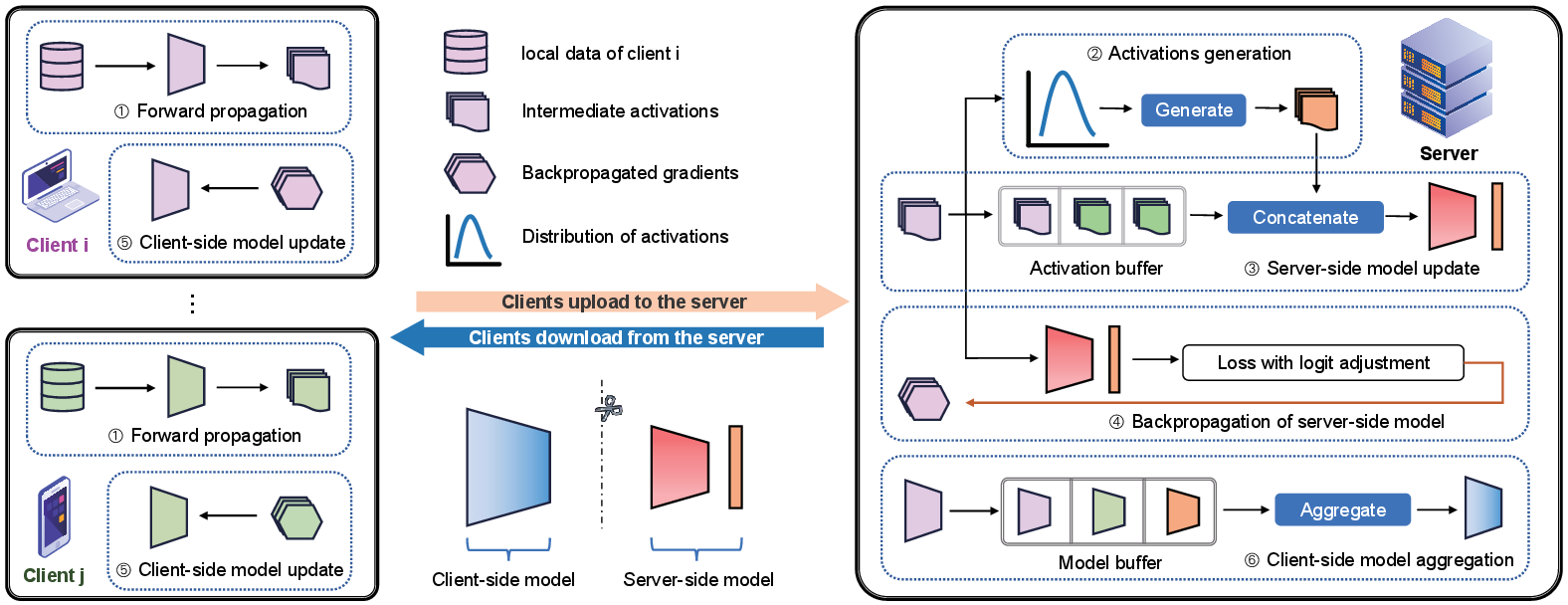} 
\caption{The framework of GAS.
The client-side model is updated through four steps: \ding{172} Clients perform forward propagation; \ding{175} The server receives the activations and computes backpropagated gradients; \ding{176} Clients receive the gradients to update the client-side models, and complete a local iteration. After finishing local iterations, clients send the updated client-side models to the server; \ding{177} The server stores these models in the model buffer and, when full, aggregates them to complete a global iteration. The server-side model is updated through two steps: \ding{173}  Received activations update the distributions of activations. When the activation buffer is full, the server generate activations from these distributions; \ding{174} Activations are stored in the buffer and, when full, the server concatenates them with generative activations to update the server-side model.
}
\label{figure1}
\end{figure*}

\subsection{Preliminaries}
Consider a SFL scenario involving $ K $ clients indexed by $ \mathcal{K} = \{1, 2, \ldots, K\} $. Each client $ k $ holds a local dataset $ \mathcal{D}_k $ with $ |\mathcal{D}_k| $ data points. The clients collaborate to train a global model $ \mathbf{w} $ under the coordination of the server. In SFL, the global model $ \mathbf{w} $ is split into two parts: the client-side model $ \mathbf{w}_c $ and the server-side model $ \mathbf{w}_s $. The clients perform local computations on $ \mathbf{w}_c $ and send the activations to the server, which completes the forward and backward passes using $ \mathbf{w}_s $. Thus the empirical loss for client $k$ is defined as
\begin{align}
    f_k(\mathbf{w};\tilde{\mathcal{D}}_k)=l(\mathbf{w}_s;h(\mathbf{w}_c;\tilde{\mathcal{D}}_k)),
\end{align}
with $ h $ representing the client-side function that maps the sampled mini-batch data $\tilde{\mathcal{D}}_k$ to the intermediate activations, and $ l $ representing the server-side function that maps activations to the final loss value. We assume partial client participation and the primary objective is to minimize the global loss function over the participating clients $\mathcal{C} $, formulated as
\begin{align}
    \min_{\mathbf{w}} F(\mathbf{w}) =  \frac{\sum_{k\in\mathcal{C}}|\mathcal{D}_k| F_k(\mathbf{w})}{\sum_{k\in\mathcal{C}}|\mathcal{D}_k|},
\end{align}
where $F_k(\mathbf{w}) $ is the local expected loss function for client $ k $ and it is unbiasedly estimated by the empirical loss $f_k(\mathbf{w};\tilde{\mathcal{D}}_k)$, such that $\mathbb{E}_{\tilde{\mathcal{D}}_k\thicksim\mathcal{D}_k}f_k(\mathbf{w},\tilde{\mathcal{D}}_k) = F_k(\mathbf{w})$.

\subsection{Overall Structure}

In Fig. \ref{figure1}, we illustrate the six key steps of GAS. The pseudo-code is illustrated in Technical Appendix A. At the beginning of the training process, the server sets the number of local iterations $E$ and global iterations $T$, with local iterations indexed by $e$ and global iterations indexed by $t$. The server then initializes an activation buffer $\mathcal{A}$ to store the received activations and their corresponding labels, and a model buffer $\mathcal{M}$ to store the received client-side models along with the respective client data sizes. Next, the server sets the minibatch size to $B$, the activation buffer size to $Q_sB$, and the model buffer size to $Q_c$.
Additionally, the server initializes the global model as $\mathbf{w}^0 = [\mathbf{w}_c^0, \mathbf{w}_s^0]$ and selects $C$ initial clients to participate in the training. A detailed description of the training process follows.
\par
\begin{itemize}
\item \textbf{Forward propagation of the client-side model (Fig. \ref{figure1}\ding{172}):} The selected client $k$ receives the client-side model and randomly selects a minibatch $\tilde{\mathcal{D}}_k$ with a batch size of $B$ from its local dataset $ \mathcal{D}_k $. The minibatch  $\tilde{\mathcal{D}}_k$  is defined as  $\tilde{\mathcal{D}}_k = \{(\mathbf{x}_1,y_1),(\mathbf{x}_2,y_2),\ldots,(\mathbf{x}_{B},y_{B})\}$, where the input samples are  $\mathbf{X}_k = \{\mathbf{x}_1,\mathbf{x}_2,\ldots,\mathbf{x}_{B}\}$  and their corresponding labels are  $\mathbf{Y}_k = \{y_1,y_2,\ldots,y_{B}\}$. The client $k$ then performs forward propagation using the client-side model $\mathbf{w}_c$  to compute the activations  $\mathbf{A}_k$  of the last layer of the client-side model, given by
\begin{align}\label{A_k}
    \mathbf{A}_k = h(\mathbf{w}_c; \tilde{\mathcal{D}}_k).
\end{align}
Upon completing the computation, the activations $\mathbf{A}_k$ along with the label set $\mathbf{Y}_k$ are sent to the server.
\item \textbf{Activations generation (Fig. \ref{figure1}\ding{173}) and  server-side model update (Fig. \ref{figure1}\ding{174}):} The server receives activations from selected client $k$ and stores them in the activation buffer as 
\begin{align}\label{store_activation}
    \mathcal{A} \leftarrow \mathcal{A} \cup (\mathbf{A}_k, \mathbf{Y}_k).
\end{align}
Additionally, the server maintains an activation distribution for each label, which is dynamically updated based on the received activations. (detailed in the next section). When the number of activations in the activation buffer exceeds $Q_sB$, the server generates activations $\widehat{\mathbf{A}}$ from the distribution. The generative activations $\widehat{\mathbf{A}}$ are then concatenated with the activations in the buffer as $\mathbf{A}_{\text{concat}} = \text{concat}(\mathbf{A}_1, \mathbf{A}_2, \ldots, \mathbf{A}_{Q_s}, \widehat{\mathbf{A}})$. Similarly, the corresponding labels $\widehat{\mathbf{Y}}$ are concatenated with the labels in the buffer as $\mathbf{Y}_{\text{concat}} = \text{concat}(\mathbf{Y}_1, \mathbf{Y}_2, \ldots, \mathbf{Y}_{Q_s}, \widehat{\mathbf{Y}})$. Finally, the concatenated activations $\mathbf{A}_{\text{concat}}$ are used as input to update the server-side model:
\begin{align}\label{update_server}
    \mathbf{w}_{s}^{e+1}=\mathbf{w}_{s}^{e}-\eta\nabla_{\mathbf{w}_{s}^{e}}l(\mathbf{w}_{s}^{e};\mathbf{A}_{\text{concat}},\mathbf{Y}_{\text{concat}}).
\end{align}
\item \textbf{Backpropagation of server-side model (Fig. \ref{figure1}\ding{175})} The server computes the backpropagated gradients based on the received activations. As logit adjustment \cite{menon2021longtail,zhang2022federated} is popular for improving model performance under conditions of data heterogeneity, we apply it to calibrate the loss function of each client, as follows:
\begin{align}
    l_k(\mathbf{w}_{s};\mathbf{A}_{k},\mathbf{Y}_{k})=-\log\left[\frac{e^{s_{y}}+\log P_{k}(y)}{\sum_{{y^{\prime}=1}}^{M}e^{s_{y^{\prime}}+\log P_{k}(y^{\prime})}}\right],
\end{align}
where $s_{y}$ is predicted score for label $y$, $P_k(y)$ is the label distribution of client $k$ and $M$ is the total number of classes.
Thus the backpropagated gradient is computed as 
\begin{align}\label{G_k}
    \mathbf{G}_k = \nabla_{\mathbf{A}_k} l_k(\mathbf{w}_s; \mathbf{A}_k, \mathbf{Y}_k),
\end{align}
which is then sent to client $k$.
\item \textbf{Backpropagation of client-side model (Fig. \ref{figure1}\ding{176}):}  The client $k$ performs backpropagation using the received gradient and updates its local client-side model using the chain rule, given by
\begin{align}\label{update_client}
    &\mathbf{w}_{c,k}^{e+1} = \mathbf{w}_{c,k}^{e} \nonumber\\
    &~~~~- \eta \nabla_{\mathbf{A}_k} l_k(\mathbf{w}_s; \mathbf{A}_k, \mathbf{Y}_k) \nabla_{\mathbf{w}_{c,k}^{e}} h_k(\mathbf{w}_{c,k}^{e}; \mathbf{X}_k).
\end{align}
When client $k$ completes $E$ local iterations, it sends the locally updated client-side model to the server. 
\item \textbf{Update of client-side model (Fig. \ref{figure1}\ding{177}):} The server receives the updated client-side model and stores it in the model buffer as 
\begin{align}\label{store_model}
    \mathcal{M} \leftarrow \mathcal{M} \cup (\mathbf{w}_{c,k}, |\mathcal{D}_k|),
\end{align}
When the number of models in the model buffer exceeds $Q_c$, the server aggregates these models as the current client-side model, given by
\begin{align}\label{aggregate_model}
\mathbf{w}_c^{\text{agg}} = \frac{ \sum_{(\mathbf{w}_{c,k}, |\mathcal{D}_k|) \in \mathcal{M}} |\mathcal{D}_k|\mathbf{w}_{c,k}}{\sum_{(\mathbf{w}_{c,k}, |\mathcal{D}_k|) \in \mathcal{M}}|\mathcal{D}_k|}
\end{align}
Then the server selects new client to participate in the training and sends it the current client-side model.
\end{itemize}
\par
Note that the server-side model and the client-side models are not updated synchronously. Since the frequency of server-side model updates and client-side model aggregations is determined by the activation buffer size and the model buffer size, we typically set $Q_s = Q_c$ to ensure consistency in model updates. Additionally, to clarify the notation, we define a global iteration as $E$ updates of the server-side model. After $E \times T$ iterations, the trained server-side model $\mathbf{w}_s^T$ is obtained. We define each aggregation of client-side models as a global iteration and after $T$ aggregations, the trained client-side model $\mathbf{w}_c^T$ is obtained. 

\subsection{Generative Activation-Aided Updates}
Due to the activations being uploaded asynchronously by selected clients, the activation buffer frequently receives activations from resource-rich clients. This results in a bias in the server-side model updates. To address this issue, we propose a method called Generative Activation-Aided Updates (Fig. \ref{figure1}\ding{173} and Fig. \ref{figure1}\ding{174}), where the server maintains the distribution of activations for each label $y$, represented as a Gaussian distribution $\mathcal{N}_y(\boldsymbol{\mu}_y, \boldsymbol{\Sigma}_y)$. The server generates activations from these distributions to assist in updating the server-side model. The key steps are as follows:

\begin{itemize}
\item \textbf{Dynamic Weighted Update:} The server dynamically updates the mean $\boldsymbol{\mu}$ and variance $\boldsymbol{\Sigma}$ of the activation distribution using asynchronously uploaded activations in a weighted manner. Specifically, we define the weighting function as $s(n)$, where $n$ represents the training progress, denoted by the total number of iterations $n=tE+e$. Since activations are uploaded asynchronously, each activation has a different training progress. We define $n(\mathbf{A})$ as the training progress of activation $\mathbf{A}$. The weighted mean for a training progress of $N$ can be expressed as: $\boldsymbol{\mu}_N=\frac1{S_N}\sum_{\mathbf{A}\in\mathcal{A}_N} s(n(\mathbf{A}))\mathbf{A}$. And the weighted variance is given by $\boldsymbol{\Sigma}_N=\frac1{S_N}\sum_{\mathbf{A}\in\mathcal{A}_N}s(n(\mathbf{A})) (\mathbf{A}-\boldsymbol{\mu}_N)(\mathbf{A}-\boldsymbol{\mu}_N)^T$, where $\mathcal{A}_N$ denotes the set of all activations uploaded to the server up to training progress $N$, and $S_N$ is the sum of the weights, defined as $S_N=\sum_{\mathbf{A}\in\mathcal{A}_N}s(n(\mathbf{A}))$. Since activations are dynamically uploaded, we adopt a dynamic update approach. Given a newly received activation $\mathbf{A}$, the mean is dynamically updated by
\begin{align}
    \boldsymbol{\mu}_N=\frac{S_{N-1}}{S_{N-1} + s(n(\mathbf{A}))}\boldsymbol{\mu}_{N-1}+\frac{s(n(\mathbf{A}))}{S_{N-1} + s(n(\mathbf{A}))}\mathbf{A}.
\end{align}
The variance is dynamically updated by
\begin{align}
    \mathbf{\Sigma}_{N}=&\frac{S_{N-1}(\boldsymbol{\Sigma}_{N-1}+(\boldsymbol{\mu}_{N}-\boldsymbol{\mu}_{N-1})(\boldsymbol{\mu}_{N}-\boldsymbol{\mu}_{N-1})^{T})}{S_{N-1} + s(n(\mathbf{A}))} \nonumber\\
    &+\frac{s(n(\mathbf{A}))(\boldsymbol{\mu}_{N}-\mathbf{A})(\boldsymbol{\mu}_{N}-\mathbf{A})^{T}}{S_{N-1} + s(n(\mathbf{A}))}.
\end{align}
The Derivation can be founded in Technical Appendix B.
\item \textbf{Generating and Concatenation:} During the server-side model update, the server generates activations $\widehat{\mathbf{A}}$ by sampling from the distributions according to the skewness of the labels. For instance, the server adjusts the sampling to ensure that each label has an equal amount of data. These generative activations are then concatenated with the activations in the activation buffer to form the input for updating the server-side model as (\ref{update_server}). This method ensures that the server-side model receives a more balanced set of activations, mitigating the bias introduced by stragglers.
\end{itemize}
Note that we consider newer activations to be more important. Therefore, we define an increasing weighting function, such as an exponential function $s(n) = ae^{bn}$ or a polynomial function $s(n) = an^b$ \cite{xie2019asynchronous,liu2024fedasmu}, where stale activations become less significant as training progresses, thereby mitigating the impact of stragglers on the activation distribution updates.

\section{Theoretical Analysis}
In this section, we provide a theoretical analysis to better understand the error bound and performance improvement of the proposed GAS. Since the server-side model and the client-side model are updated independently, where the parameters of one model remain fixed while the other is updated, we separately analyze the convergence rates of the server-side model and the client-side model. To ensure clarity, we denote $f_k(\mathbf{w}_c)$ as the local loss function of the client-side model $h(\mathbf{w}_c;\tilde{\mathcal{D}}_k)$, and $f_s(\mathbf{w}_s)$ as the loss function of the server-side model $l(\mathbf{w}_{s};\mathbf{A}_{\text{concat}},\mathbf{Y}_{\text{concat}})$.
Our analysis is based on the following assumptions:
\begin{assumption}\label{assumption1}
(Smoothness) Loss function of server-side model and each local loss function of client-side model are Lipschitz smooth, i.e., for all $\mathbf{w}$ and $\mathbf{w}^{\prime}$, $\|\nabla_{\mathbf{w}_{s}} f_s(\mathbf{w}_{s}) - \nabla_{\mathbf{w}_{s}} f_s(\mathbf{w}_{s}^{\prime})\| \leq \gamma_1 \|\mathbf{w}_{s} - \mathbf{w}_{s}^{\prime}\|$ and  $\|\nabla_{\mathbf{w}_{c}} f_k(\mathbf{w}_{c}) - \nabla_{\mathbf{w}_{c}} f_k(\mathbf{w}_{c}^{\prime})\| \leq \gamma_2 \|\mathbf{w}_{c} - \mathbf{w}_{c}^{\prime}\|$.
\end{assumption}
\begin{assumption}\label{assumption2}
(Bounded Gradient Variance) The stochastic gradient of server-side model $\nabla_{\mathbf{w}_{s}} f_s(\mathbf{w}_{s})$ and the stochastic gradient of client-side model $\nabla_{\mathbf{w}_{c}} f_k(\mathbf{w}_{c})$ have bounded variance: $\mathbb{E}[\|\nabla_{\mathbf{w}_s}f_s(\mathbf{w}_s)-\nabla_{\mathbf{w}_s}F_s(\mathbf{w}_s)\|^2]\leq\frac{\sigma^2}{BQ_s}$ and $\mathbb{E}[\|\nabla_{\mathbf{w}_c}f_k(\mathbf{w}_c)-\nabla_{\mathbf{w}_c}F_k(\mathbf{w}_c)\|^2]\leq\frac{\sigma^2}B$.
\end{assumption}
\begin{assumption}\label{assumption3}
(Bounded Dissimilarity) 
In server-side model updates, gradient dissimilarity is referred to as the bias caused by stragglers, which is bounded as:
$\mathbb{E}\left[\|\nabla_{\mathbf{w}_s}F_s(\mathbf{w}_s)-\nabla_{\mathbf{w}_s}F(\mathbf{w}_s)\|^2\right]\leq\kappa_1^2$.
In client-side model updates, gradient dissimilarity is referred to as the bias caused by data heterogeneity across clients, which is bounded as: 
$\mathbb{E}\left[||\nabla_{\mathbf{w}_c}F_k(\mathbf{w}_c)-\nabla_{\mathbf{w}_c}F(\mathbf{w}_c)||^2\right]\leq\kappa_2^2$. 
\end{assumption}
In the proposed GAS, the activation distribution gradually approximates the ground-truth activation distribution through dynamic updates. This leads us to the following lemma:
\begin{lemma}\label{lemma1}
By introducing generative activations, the server-side model update achieves a tighter bounded dissimilarity, as shown below:
\begin{align}
&\mathbb{E}_{\substack{(\mathbf{A}, \mathbf{Y}) \sim \mathcal{A} \\ \widehat{\mathbf{A}} \sim \mathcal{N}}} \left[\left\|\nabla_{\mathbf{w}_s}F_s(\mathbf{w}_s)-\nabla_{\mathbf{w}_s}F(\mathbf{w}_s)\right\|^2\right] \nonumber \\
&~~~~~\leq\mathbb{E}_{(\mathbf{A},\mathbf{Y})\sim\mathcal{A}}\left[\left\|\nabla_{\mathbf{w}_s}F_s(\mathbf{w}_s)-\nabla_{\mathbf{w}_s}F(\mathbf{w}_s)\right\|^2\right].
\end{align}
The Proof can be founded in Technical Appendix C.
\end{lemma}
This reduction in gradient dissimilarity indicates that the server-side model update becomes less biased by concatenating generative activations. Now, we are ready to state the following theorem, which provides the convergence upper bounds for the proposed GAS, considering both the client-side model and the server-side model.
\begin{theorem}
When Assumptions \ref{assumption1}-\ref{assumption3} hold, given the learning rate $\eta \leq \frac{1}{\gamma_1}$, the convergence rate of server-side model is given by
\begin{align}\label{t1eq1}
\frac{1}{ET} &\sum_{t=0}^{T-1} \sum_{e=0}^{E-1} \mathbb{E} \left[ \|\nabla_{\mathbf{w}_s} F(\mathbf{w}_s^{t,e})\|^2 \right] \nonumber\\
&\leq \mathcal{O} \left( \frac{F(\mathbf{w}_s^0) - F^*}{ET \eta } + \frac{\eta \sigma^2}{BQ_s + \widehat{B}} + \kappa^2_1 \right),
\end{align}
where $\widehat{B}$ is the batch size of generative activations. 
\par
Given the learning rate $\eta \leq \frac{1}{20 \gamma_2 \sqrt{\tau_{\max}}}$ and the maximum upload delay of client-side model  $\tau_{\max}$, the convergence rate of client-side model is given by
\begin{align}\label{t1eq2}
&\frac{1}{T} \sum_{t=0}^{T-1}  \mathbb{E} \left[ \|\nabla_{\mathbf{w}_c} F(\mathbf{w}_c^t)\|^2 \right] 
\leq \mathcal{O} \Bigg( \frac{F(\mathbf{w}_s^0) - F^*}{ET \eta } \nonumber \\
&~~~~~~+ \left( \frac{\sigma^2}{B} + \kappa_2^2 \right) \eta E + \left( \frac{\sigma^2}{B} + \kappa_2^2 \right) \eta^2 E^2 \tau_{\max}^2 \Bigg).
\end{align}
The Proof can be founded in Technical Appendix D.
\end{theorem}
From \eqref{t1eq1}, it is evident that stragglers primarily affect the convergence performance through their impact on the bounded dissimilarity of the server-side model $\kappa^2_1$. Specifically, if there is a bias in the activations stored in the activation buffer, the bounded dissimilarity increases, leading to an increase in $\kappa^2_1$. This, in turn, enlarges the convergence upper bound in \eqref{t1eq1}. According to Lemma \ref{lemma1}, the proposed method achieves a tighter bounded dissimilarity by introducing generative activations. As a result, the server-side model attains a tighter upper bound, enhancing convergence performance.
From \eqref{t1eq2}, it is evident that stragglers primarily affect convergence performance through $\tau^2_{\max}$, which is multiplied by the learning rate $\eta$. By setting a learning rate that decays over the global iterations, i.e., $\eta^t = \eta^0 / \sqrt{t}$, the impact of $\tau^2_{\max}$ will be gradually mitigated as the training progresses.

\section{Experiments}
\subsection{Implementation Details}
Unless otherwise stated, the number of clients is set to $20$, with $10$ clients participating in each global iteration. Each client performs $20$ local iterations with a learning rate of $0.01$ and a minibatch size of $32$. We use a linearly increasing weighting function, i.e., $s(n) = n$  and select AlexNet as the model architecture, where we set up the first $6$ layers as the client-side model and the last $8$ layers as the server-side model. 
To simulate a real-word communication environment, we consider a cell network with a radius of $1000$ meters. The server is placed at the center of the network, with clients randomly and uniformly distributed within the cell. The path loss between each client and the server is modeled as $128.1 + 37.6 \log_{10}(r)$ dB, where $r$ is the distance from the client to the server in kilometers, according to \cite{abetaevolved}. The client transmit power is uniformly set to $0.2$ W. We assume orthogonal uplink channel access with a total bandwidth $W = 10$ MHz and a power spectrum density of the additive Gaussian noise $N_0 = -174$ dBm/Hz. Additionally, clients are assigned random computational capabilities, ranging between $10^9$ and $10^{10}$ FLOPs. More experimental details can be founded in Technical Appendix E.
\subsection{Baseline Settings}
For the baseline comparison, we include both asynchronous and synchronous FL algorithms. The baseline asynchronous FL algorithms are FedBuff \cite{nguyen2022federated} and CA$^2$FL \cite{wang2023tackling}. FedBuff introduces a buffer mechanism to enable asynchronous FL, while CA$^2$FL  builds on FedBuff by incorporating cached update calibration to enhance model performance in the presence of client data heterogeneity. Additionally, we select MiniBatch-SFL \cite{huang2023minibatch} and S$^2$FL \cite{yan2023have} as baseline synchronous SFL algorithms. MiniBatch-SFL improves SFL performance by updating server-side model centrally, while S$^2$FL builds on MiniBatch-SFL by introducing adaptive model splitting and activation grouping strategies to address the stragglers issue.
\subsection{Dataset Settings}
The datasets used for evaluation include CIFAR-10 \cite{krizhevsky2009learning}, CINIC-10 \cite{darlow2018cinic}, and Fashion-MNIST \cite{xiao2017fashion}. To simulate data heterogeneous, we employ both shard-based and distribution-based label skew methods \cite{zhang2022federated}. The shard-based method involves sorting data by labels and dividing it into multiple shards. Each client receives a subset of these shards, resulting in training data with only a few labels for each client. We denote the data heterogeneity of this method by $\text{shard}$, where $\text{shard}=2$ indicates each client has at most $2$ types of data. This method represents an extreme form of data heterogeneity. In addition, the distribution-based label skew method allocates data to clients based on a Dirichlet distribution. Each client receives a proportion of samples from each label according to this distribution, resulting in a mix of majority and minority classes, and potentially some missing classes. We denote the data heterogeneity of this method by $\alpha$, where $\text{Dir}(\alpha)$ indicates the Dirichlet distribution. The smaller the value of $\alpha$, the higher the degree of data heterogeneity. This method better reflects real-world data heterogeneity.
\subsection{Validation of Theoretical Analysis}
\begin{figure}[t]
  \centering 
  \subfigure[Gradient dissimilarity.]{ 
    \includegraphics[clip, viewport= 10 0 530 400,width=3.9cm]{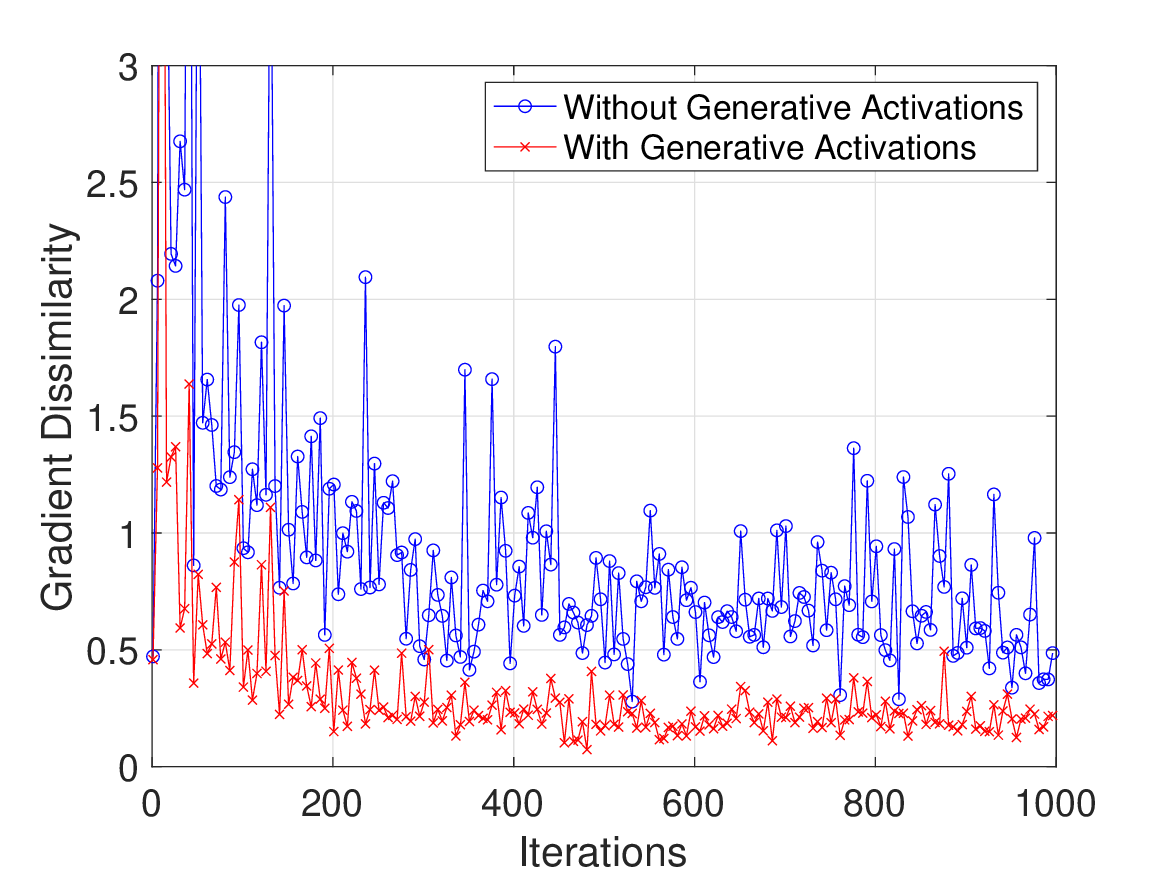} 
  } 
  \subfigure[Test accuracy.]{ 
    \includegraphics[clip, viewport= 10 0 530 400,width=3.9cm]{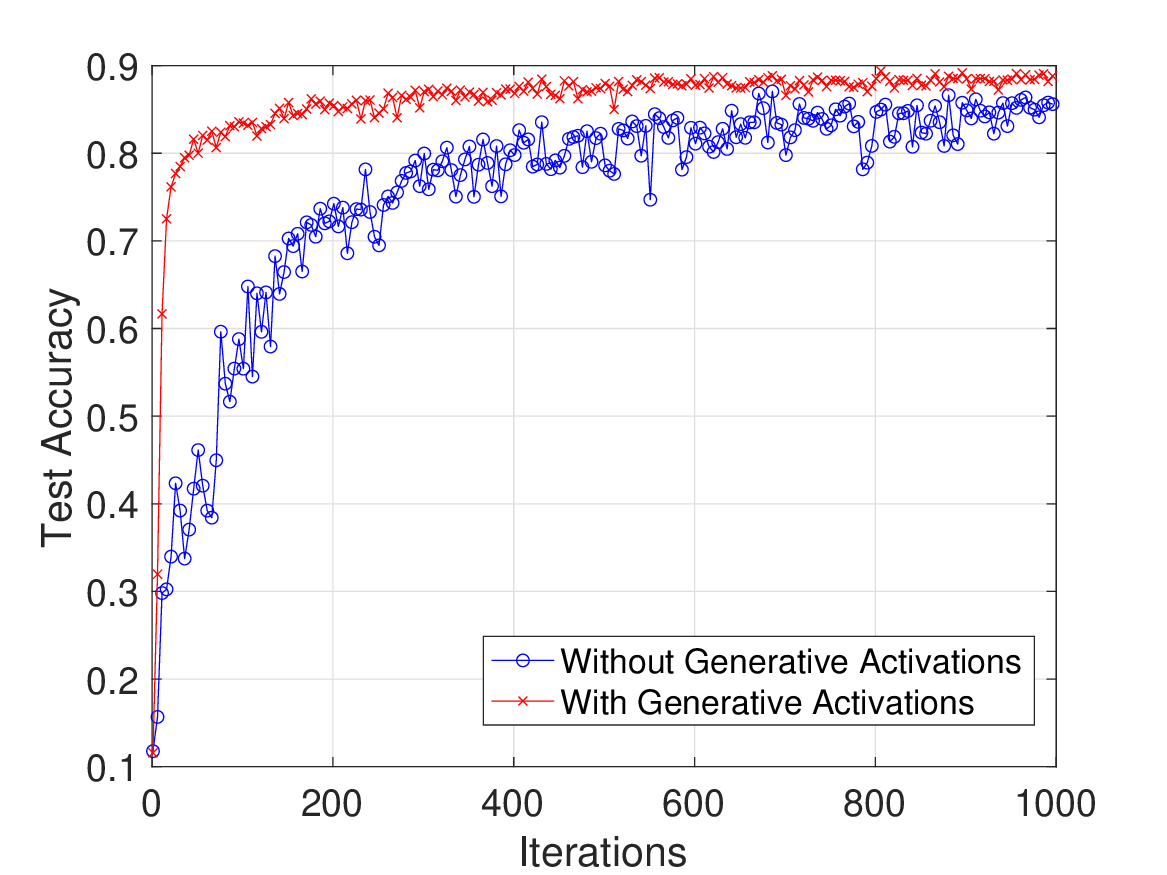} 
  } 
  \caption{Impact of generative activations on gradient dissimilarity and convergence performance.} 
\label{figure2}
\end{figure}

In this subsection, we validate our theoretical analysis by assessing gradient dissimilarity both without and with generative activations using the Fashion-MNIST datasets under heterogeneity conditions with $\text{shard}=2$. The total number of clients is set to $10$, with $3$ clients participating in each global iteration. The experimental results are depicted in Fig. \ref{figure2}.
As shown in Fig. \ref{figure2} (a), the gradient dissimilarity is reduced by introducing generative activations, thereby confirming Lemma 1. This result demonstrates that the proposed method can achieve tighter bounded dissimilarity via the use of generative activations.
Fig. \ref{figure2} (b) further illustrates that the introduction of generative activations enhances convergence speed and achieve better accuracy, thus confirming Theorem 1. This indicates that tighter bounded dissimilarity reduces the upper bound of convergence rate, leading to superior convergence performance.
\subsection{Performance Evaluation}
\begin{table*}[t]
\centering
\resizebox{\linewidth}{!}{
\begin{tabular}{ccccccccc}
\toprule
\multirow{2}{*}{Method} & \multicolumn{2}{c}{CIFAR-10}             & \multicolumn{4}{c}{CINIC-10}                                                       & \multicolumn{2}{c}{Fashion-MNIST}            \\ 
\cmidrule(lr){2-3}\cmidrule(lr){4-7}\cmidrule(lr){8-9}
                        & $\text{s}=2$   & $\alpha=0.1$       & $\text{s}=2$   & $\text{s}=4$   & $\alpha=0.1$       & $\alpha=0.3$       & $\text{s}=2$   & $\alpha=0.1$       \\ \midrule
FedAvg                  & $72.88_{\pm 5.71}$ & $70.49_{\pm 4.24}$ & $52.66_{\pm 6.51}$ & $62.26_{\pm 2.52}$ & $57.17_{\pm 1.04}$ & $65.46_{\pm 2.09}$ & $87.99_{\pm 2.12}$ & $88.74_{\pm 1.19}$ \\ 
FedBuff                 & $69.04_{\pm 3.51}$ & $71.82_{\pm 2.86}$ & $48.98_{\pm 0.87}$ & $58.32_{\pm 2.20}$ & $54.23_{\pm 2.11}$ & $64.69_{\pm 1.81}$ & $84.93_{\pm 4.11}$ & $85.81_{\pm 3.68}$ \\
CA$^2$FL                & $79.57_{\pm 0.98}$ & $78.56_{\pm 0.99}$ & $64.29_{\pm 1.45}$ & $68.42_{\pm 1.11}$ & $64.27_{\pm 0.78}$ & $68.77_{\pm 0.92}$ & $88.32_{\pm 1.19}$ & $89.07_{\pm 0.58}$ \\ \midrule
Ours                    & $\mathbf{82.78}_{\pm 0.58}$ & $\mathbf{81.72}_{\pm 0.50}$ & $\mathbf{68.32}_{\pm 0.17}$ & $\mathbf{70.29}_{\pm 0.27}$ & $\mathbf{65.94}_{\pm 1.14}$ & $\mathbf{69.36}_{\pm 0.65}$ & $\mathbf{90.66}_{\pm 0.20}$ & $\mathbf{90.58}_{\pm 0.34}$ \\  \bottomrule
\end{tabular}}
\caption{Test accuracy (\%) on CIFAR-10, CINIC-10 and Fashion-MNIST.}
\label{table1}
\end{table*}

\begin{figure}[t]
  \centering 
  \subfigure[CIFAR-10 with $\text{shard}=2$.]{ 
    \includegraphics[clip, viewport= 10 0 530 400,width=3.9cm]{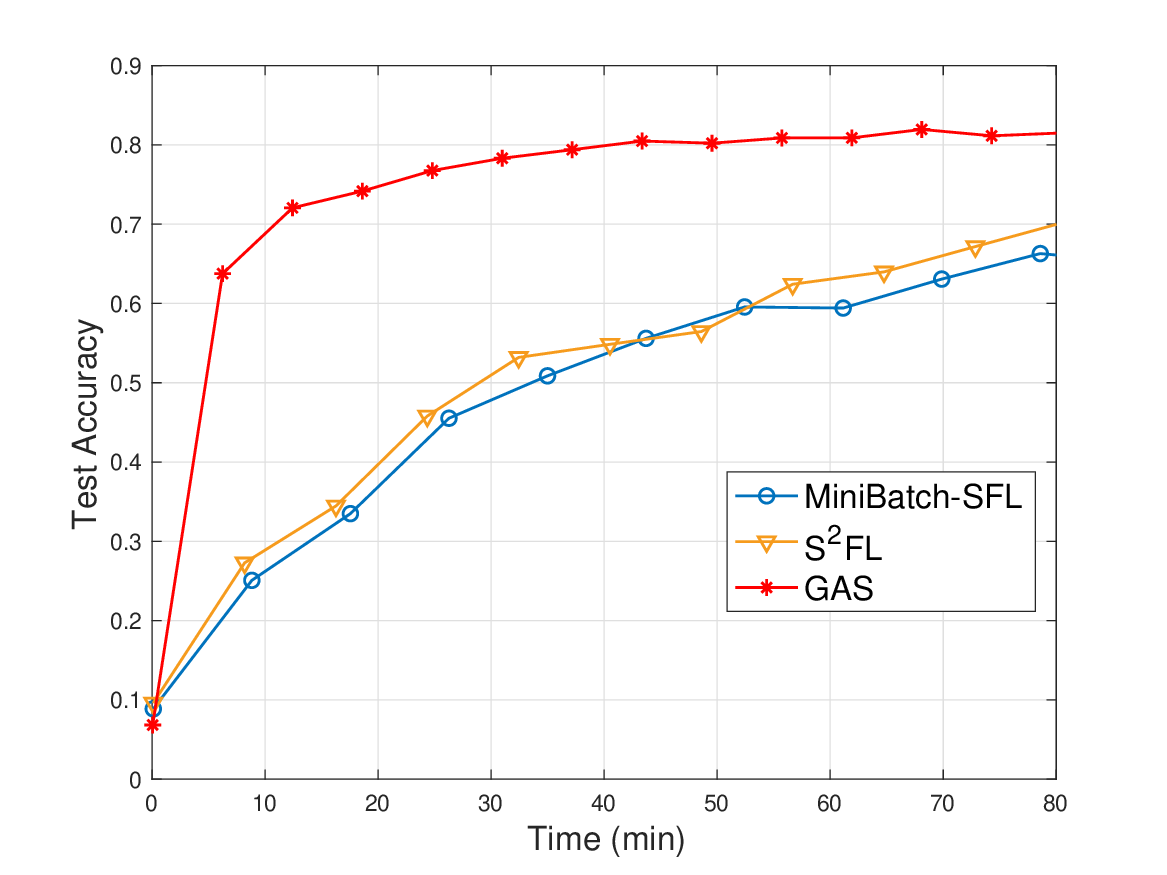} 
  } 
  \subfigure[CINIC-10 with $\text{shard}=2$.]{ 
    \includegraphics[clip, viewport= 10 0 530 400,width=3.9cm]{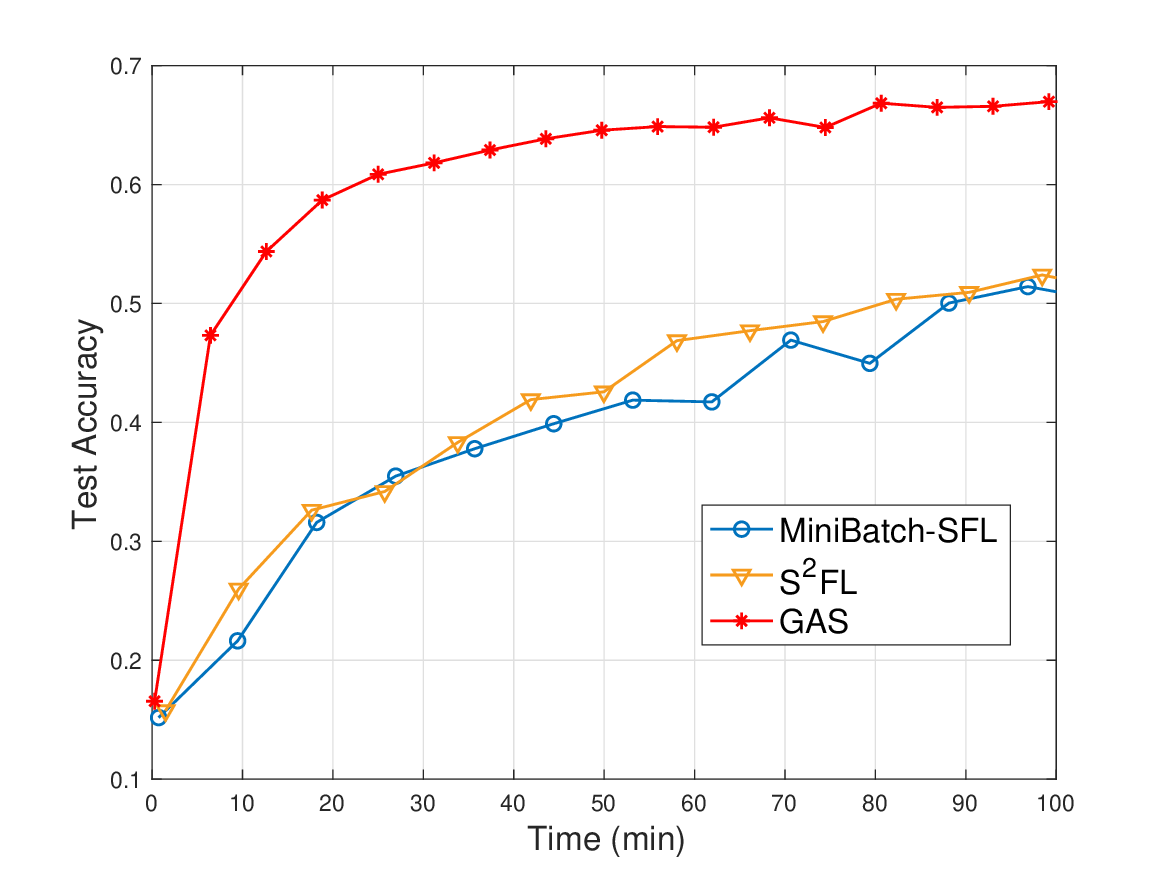} 
  } 
  \caption{Test accuracy of GAS compared with the baseline methods on CIFAR-10 and CINIC-10.} 
\label{figure3}
\end{figure}

\begin{table*}[t]
\centering
\resizebox{\linewidth}{!}{
\begin{tabular}{ccccccccc}
\toprule
\multirow{2}{*}{Method} & \multicolumn{4}{c}{$\text{shard}=2$}                                                                                  & \multicolumn{4}{c}{$\alpha=0.1$}                                                                                      \\ \cmidrule(lr){2-5}\cmidrule(lr){6-9}

                        & $E=10$                      & $E=20$                      & $E=35$                      & $E=50$                      & $E=10$                      & $E=20$                      & $E=35$                      & $E=50$                      \\ \midrule
FedBuff                 & $41.52_{\pm 2.48}$          & $40.07_{\pm 1.42}$          & $44.80_{\pm 4.09}$          & $47.84_{\pm 5.69}$          & $44.87_{\pm 3.75}$          & $51.18_{\pm 1.74}$          & $54.03_{\pm 5.01}$          & $55.77_{\pm 3.58}$          \\
CA$^2$FL                & $58.77_{\pm 0.52}$          & $62.12_{\pm 1.39}$          & $63.14_{\pm 0.28}$          & $63.31_{\pm 1.18}$          & $58.39_{\pm 1.15}$          & $60.95_{\pm 0.84}$          & $62.02_{\pm 2.39}$          & $62.73_{\pm 1.16}$          \\ \midrule
Ours                    & $\mathbf{63.07}_{\pm 0.08}$ & $\mathbf{65.58}_{\pm 0.71}$ & $\mathbf{65.09}_{\pm 1.10}$ & $\mathbf{62.40}_{\pm 1.97}$ & $\mathbf{60.68}_{\pm 1.13}$ & $\mathbf{63.39}_{\pm 2.01}$ & $\mathbf{63.12}_{\pm 2.29}$ & $\mathbf{61.96}_{\pm 3.50}$ \\ \bottomrule
\end{tabular}}
\caption{Test accuracy (\%) under different number of local iterations.}
\label{table2}
\end{table*}

In this subsection, we evaluate the performance of the proposed method across different datasets and varying degrees of data heterogeneity. For Fashion-MNIST, CIFAR-10, and CINIC-10, we employ $1000$, $2000$, and $2000$ global iterations.
We first compare our method with baseline asynchronous FL algorithms. Each experiment is run with three random seeds, and the average accuracy and standard deviation are reported in Table \ref{table1}. The experimental results demonstrate that the proposed method outperforms the baseline methods, particularly under conditions of extreme data heterogeneity. This improvement in model accuracy can be attributed to two key factors. First, the proposed method allows for centralized updates of the server-side model, significantly mitigating the issue of deep model drift caused by data heterogeneity \cite{luo2021no}. Second, by introducing generative activations, the proposed method alleviates the server-side model update bias introduced by stragglers, further enhancing model performance.
Additionally, we compare our method with baseline synchronous SFL algorithms in a real-world communication environment, with results shown in Fig. \ref{figure3}. The experimental results indicate that the proposed method exhibits better convergence performance compared to baseline methods. This improvement is due to the asynchronous transmissions of activations and client-side models, which substantially reduce training time and achieves faster convergence speeds.

\subsection{Ablation Study on Local Iterations}

\begin{figure}[t]
  \centering 

  \subfigure[$E=20$ and $\alpha=0.1$.]{ 
    \includegraphics[clip, viewport= 10 0 530 400,width=3.9cm]{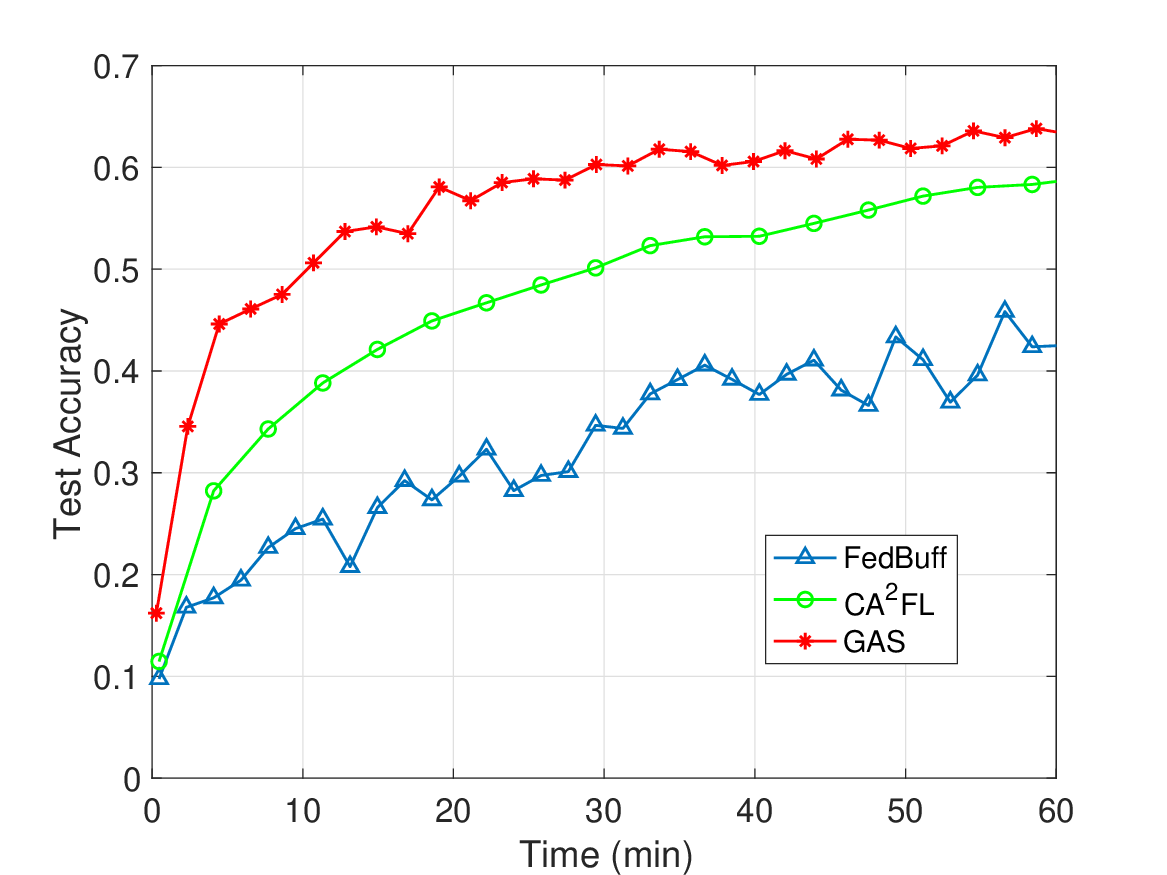} 
  } 
  \subfigure[$E=50$ and $\alpha=0.1$.]{ 
    \includegraphics[clip, viewport= 10 0 530 400,width=3.9cm]{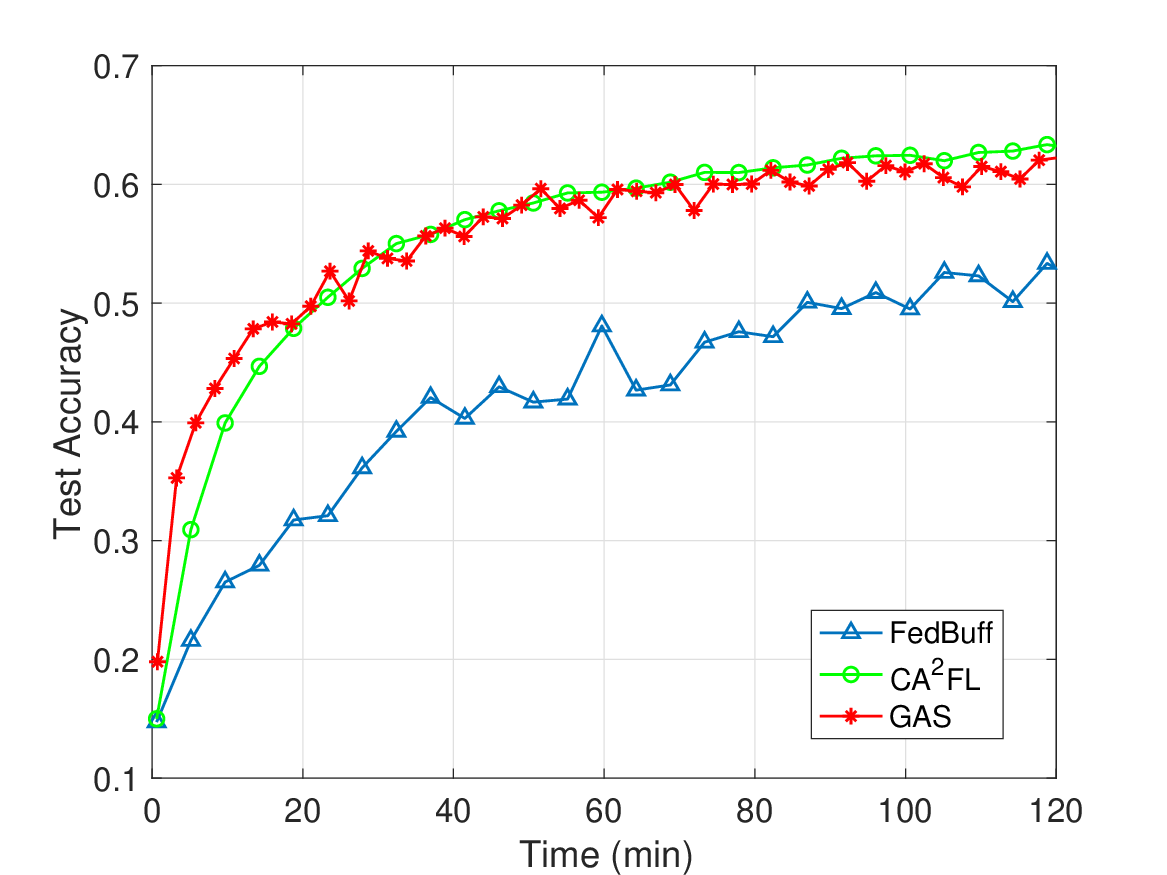} 
  } 
  \caption{Impact of local iterations on the performance of GAS compared to baseline methods.} 
\label{figure4}
\end{figure}

In this subsection, we conduct an ablation study on the number of local iterations. Unlike FL frameworks, GAS requires the additional transmissions of activations, which are influenced by the number of local iterations. Therefore, we study the impact of different local iteration settings under the real-world communication environment. We conduct experiments with local iteration settings of $10$, $20$, $35$ and $50$, while fixing the number of global iterations at $1000$. The results are presented in Table \ref{table2} and Fig. \ref{figure4}. 
As shown in Table \ref{table2}, the accuracy of GAS increases with the number of local iterations initially but decreases thereafter. This indicates that the number of local iterations must be carefully chosen to balance model accuracy and communication load. On one hand, a higher number of local iterations is necessary for sufficient local training. On the other hand, setting the number of local iterations too high can lead to local optima and increased communication load. Additionally, we observe that the accuracy of the baseline methods increase with the number of local iterations. This suggests that the baseline methods do not achieve sufficient training within the given local iteration settings. Even with a local iteration setting of $50$, the accuracy of CA$^2$FL remains lower than that of GAS with a local iteration setting of $20$, indicating the higher training efficiency of GAS.
\par
From Fig. \ref{figure4}, it is evident that GAS demonstrates faster convergence and higher accuracy compared to the baseline methods at the lower local iteration setting ($E=20$). This highlights the significant advantage of GAS in real-world communication environments. Note that although CA$^2$FL performs well with $E=50$, it incurs higher computational load due to the increased number of local iterations. Specifically, CA$^2$FL takes $60$ minutes to achieve $60\%$ accuracy, whereas GAS with $E=20$ achieves the same accuracy in just $30$ minutes.

\section{Conclusion}
In this paper, we proposed GAS (Generative activation-aided Asynchronous SFL), a distributed asynchronous learning framework designed to address the stragglers issue in SFL. By employing an activation buffer and a model buffer, along with generative activation-aided updates, GAS effectively mitigated the impact of stragglers and improved model convergence. Our theoretical analysis and experimental results demonstrated that GAS achieved higher accuracy and faster convergence compared to baseline FL and SFL methods.
\par
\noindent\textbf{Limitations:} Like other SL and SFL algorithms, GAS requires the transmission of labels and activations, which poses a risk of privacy leaks. Incorporating  privacy-preserving mechanisms of SFL \cite{xiao2021mixing,li2022ressfl} into GAS to enhance data security and broaden its applicability is a promising direction for future work.

\newpage
\appendix
\section{Technical Appendix}
\subsection{A. Pseudocode of the Proposed GAS}
\begin{algorithm}[ht]
\caption{GAS-server}
\label{Algorithm_server}
\begin{algorithmic}[1]
\STATE {\bfseries Input:} Activation cache size $Q_sB$, model cache size $Q_c$, client set $\mathcal{K}=\{1,\cdots, K\}$
\STATE {\bfseries Output:} Trained global model
\STATE {\bfseries Initialize:} $q_s=0$, $q_c=0$ and select $C$ clients to run local updates
\REPEAT
    \IF {receive activations of client $k$}
        \STATE Store activations as (4) and set $q_s \leftarrow q_s + B$
        \STATE Distributions of activations update 
        \IF{$q_s=Q_sB$}
            \STATE Activations generation 
            \STATE Update server-side model as (5) and set $q_s \leftarrow 0$    
        \ENDIF
        \STATE Compute backpropagated gradient $\mathbf{G}_k$ as (7)
        \STATE Send backpropagated gradient $\mathbf{G}_k$ to client $k$ 
    \ENDIF
    \IF{receive client-side model of client $k$}
    \STATE Store client-side model as (9) and set $q_c \leftarrow q_c +1$
    \IF{$q_c=Q_c$}
            \STATE Aggregate client-side models as (10) and set $q_c \leftarrow 0$    
    \ENDIF
    \STATE Select another client $j$ from client set $\mathcal{K}$
    \STATE Send the current client-side model to client $j$
    \ENDIF
\UNTIL convergence
\end{algorithmic}
\end{algorithm}

\begin{algorithm}[t]
\caption{GAS-client}
\label{Algorithm_client}
\begin{algorithmic}[1]
\STATE {\bfseries Input:} Number of local iterations $E$, local data $\mathcal{D}_k$, 
\STATE {\bfseries Output:} Client-side model $\mathbf{w}^E_{c,k}$
\STATE Receive the current client-side model $\mathbf{w}_c$ from the server
\FOR{$e=1,\cdots$, $E$}
    \STATE Sample a minibatch $\tilde{\mathcal{D}}_k \in \mathcal{D}_k$ and compute activations as (3)
    \STATE Upload activations $\mathbf{A}_k$ to the server
    \STATE Receive backpropagated gradient $\mathbf{G}_k$ from the server
    \STATE Perform backpropagation and update client-side model as (8)
\ENDFOR
\STATE Upload client-side model $\mathbf{w}^E_{c,k}$ to the server
\end{algorithmic}
\end{algorithm}

In this section, we present the pseudocode for the server and client modules of the proposed GAS in Algorithm \ref{Algorithm_server} and Algorithm \ref{Algorithm_client}.

\subsection{B. Derivation of Dynamic Weighted Update}
Given a newly received activation $\mathbf{A}$, the weighted mean for a training progress of $N$ can be expressed as:
\begin{align}
    &\boldsymbol{\mu}_N=\frac{1}{S_N}\sum_{\mathbf{A}^{\prime}\in\mathcal{A}_N}s(n(\mathbf{A}^{\prime}))\mathbf{A}^{\prime}\nonumber \\
&=\frac1{S_N}\Bigg(\sum_{\mathbf{A}^{\prime}\in\mathcal{A}_{N-1}}s(n(\mathbf{A}^{\prime}))\mathbf{A}^{\prime}+s(n(\mathbf{A}))\mathbf{A}\Bigg)\nonumber \\
&=\frac{S_{N-1}}{S_N}\boldsymbol{\mu}_{N-1}+\frac{s(n(\mathbf{A}))}{S_N}\mathbf{A},
\end{align}
which completes the proof.
\par
Given a newly received activation $\mathbf{A}$, the weighted variance for a training progress of $N$ can be expressed as:
\begin{align}\label{sum_variance}
    &\boldsymbol{\Sigma}_N=\frac1{S_N}\sum_{\mathbf{A}^{\prime}\in\mathcal{A}_N}s(n(\mathbf{A}^{\prime}))(\mathbf{A}^{\prime}-\boldsymbol{\mu}_N)(\mathbf{A}^{\prime}-\boldsymbol{\mu}_N)^T\nonumber \\
&=\frac1{S_N}\Bigg(\sum_{\mathbf{A}^{\prime}\in\mathcal{A}_{N-1}}s(n(\mathbf{A}^{\prime}))(\mathbf{A}^{\prime}-\boldsymbol{\mu}_N)(\mathbf{A}^{\prime}-\boldsymbol{\mu}_N)^T\nonumber \\
&\quad\quad+s(n(\mathbf{A}))(\mathbf{A}-\boldsymbol{\mu}_N)(\mathbf{A}-\boldsymbol{\mu}_N)^T\Bigg).
\end{align}
For $\sum_{\mathbf{A}^{\prime}\in\mathcal{A}_{N-1}}s(n(\mathbf{A}^{\prime}))(\mathbf{A}^{\prime}-\boldsymbol{\mu}_N)(\mathbf{A}^{\prime}-\boldsymbol{\mu}_N)^T$, we have
\begin{align}\label{sum_N_1}
    &\sum_{\mathbf{A}^{\prime}\in\mathcal{A}_{N-1}}s(n(\mathbf{A}^{\prime}))(\mathbf{A}^{\prime}-\boldsymbol{\mu}_N)(\mathbf{A}^{\prime}-\boldsymbol{\mu}_N)^T\nonumber \\
&=\sum_{\mathbf{A'}\in\mathcal{A}_{N-1}}\Big(s(n(\mathbf{A'}))(\mathbf{A'}-\boldsymbol{\mu}_{N-1}-(\boldsymbol{\mu}_N-\boldsymbol{\mu}_{N-1}))\nonumber \\
&\quad\quad(\mathbf{A}'-\boldsymbol{\mu}_{N-1}-(\boldsymbol{\mu}_N-\boldsymbol{\mu}_{N-1}))^T\Big)\nonumber \\
&=\sum_{\mathbf{A}^{\prime}\in\mathcal{A}_{N-1}}\Big(s(n(\mathbf{A}^{\prime}))((\mathbf{A}^{\prime}-\boldsymbol{\mu}_{N-1})(\mathbf{A}^{\prime}-\boldsymbol{\mu}_{N-1})^T\nonumber \\
&\quad\quad-2(\mathbf{A}^{\prime}-\boldsymbol{\mu}_{N-1})(\boldsymbol{\mu}_N-\boldsymbol{\mu}_{N-1})^T\nonumber \\
&\quad\quad+(\boldsymbol{\mu}_N-\boldsymbol{\mu}_{N-1})(\boldsymbol{\mu}_N-\boldsymbol{\mu}_{N-1})^T)\Big)\nonumber \\
&=S_{N-1}(\boldsymbol{\Sigma}_{N-1}+(\boldsymbol\mu_N-\boldsymbol\mu_{N-1})(\boldsymbol\mu_N-\boldsymbol\mu_{N-1})^T).
\end{align}
Incorporating \eqref{sum_N_1} into \eqref{sum_variance}, we can obtain
\begin{align}
        \mathbf{\Sigma}_{N}=&\frac{S_{N-1}(\boldsymbol{\Sigma}_{N-1}+(\boldsymbol{\mu}_{N}-\boldsymbol{\mu}_{N-1})(\boldsymbol{\mu}_{N}-\boldsymbol{\mu}_{N-1})^{T})}{S_{N}} \nonumber\\
    &+\frac{s(n(\mathbf{A}))(\boldsymbol{\mu}_{N}-\mathbf{A})(\boldsymbol{\mu}_{N}-\mathbf{A})^{T}}{S_{N}},
\end{align}
which completes the proof.

\subsection{C. Proof of Lemma 1}
When introducing the generative activations, gradient dissimilarity is bounded as
\begin{align}
    &\mathbb{E}_{\substack{(\mathbf{A}, \mathbf{Y}) \sim \mathcal{A} \\ \widehat{\mathbf{A}} \sim \mathcal{N}}} \left[\left\|\nabla_{\mathbf{w}_s}F_s(\mathbf{w}_s)-\nabla_{\mathbf{w}_s}F(\mathbf{w}_s)\right\|^2\right] \nonumber \\
    &=\mathbb{E}_{\substack{(\mathbf{A}, \mathbf{Y}) \sim \mathcal{A} \\ \widehat{\mathbf{A}} \sim \mathcal{N}}} \left[\left\|\frac B{BQ_s+\widehat{B}}\nabla_{\mathbf{w}_s}F_s(\mathbf{w}_s;\boldsymbol{x},y)\right.\right.\nonumber \\
    &\left.\left.+\frac{\widehat{B}}{BQ_s+\widehat{B}}\nabla_{\mathbf{w}_s}F_s(\mathbf{w}_s;\widehat{\boldsymbol{x}},y)-\nabla_{\mathbf{w}_s}F(\mathbf{w}_s)\right\|^2\right].
\end{align}
Let $g_1=\nabla_{\mathbf{w}_s}F_s(\mathbf{w}_s;\boldsymbol{x},y)$, $g_2=\nabla_{\mathbf{w}_s}F_s(\mathbf{w}_s;\widehat{\boldsymbol{x}},y)$ and $g=\nabla_{\mathbf{w}_s}F(\mathbf{w}_s)$, the original term can be written as
\begin{align}\label{g1g2g}
    \mathbb{E}\left[\left\|\frac{BQ_s}{BQ_s+\widehat{B}}(g_1-g)+\frac{\widehat{B}}{BQ_s+\widehat{B}}(g_2-g)\right\|^2\right].
\end{align}
Since $g_1$ and $g_2$ is i.i.d. and the activation distribution approximates the ground-truth activation distribution, the expectation of the cross term of (\ref{g1g2g}) is zero:
\begin{align}
    \mathbb{E}[(g_1-g)(g_2-g)]=\mathbb{E}[g_1-g]\mathbb{E}[g_2-g]=0,
\end{align}
and the gradient variance of $g_2$ satisfies
\begin{align}
    \mathbb{E}[\parallel g_2-g\parallel^2]\leq\mathbb{E}[\parallel g_1-g\parallel^2].
\end{align}
Thus, the gradient dissimilarity is bounded by
\begin{align}
    \left(\left(\frac{BQ_s}{BQ_s+\widehat{B}}\right)^2+\left(\frac{\widehat{B}}{BQ_s+\widehat{B}}\right)^2\right)\mathbb{E}[\parallel g_1-g\parallel^2].
\end{align}
Since $\left(\left(\frac{BQ_s}{BQ_s+\widehat{B}}\right)^2+\left(\frac{\widehat{B}}{BQ_s+\widehat{B}}\right)^2\right)\leq1$,  we can obtain
\begin{align}
    \mathbb{E}&\left[\left\|\frac B{BQ_s+\widehat{B}}(g_1-g)+\frac{\widehat{B}}{BQ_s+\widehat{B}}(g_2-g)\right\|^2\right]\nonumber \\
    &\leq\mathbb{E}[\parallel g_1-g\parallel^2],
\end{align}
which completes the proof.

\subsection{D. Proof of Theorem 1:}
\subsubsection{Convergence rate of server-side model:} 
When Assumption 1 holds, the decrease of the loss function can be bounded as 
\begin{align}
    \mathbb{E}&[F(\mathbf{w}_s^{t,e+1})]-F(\mathbf{w}_s^{t,e}) \nonumber \\
    &\leq\mathbb{E}[\langle\nabla_{\mathbf{w}_s}F(\mathbf{w}_s^{t,e}),\mathbf{w}_s^{t,e+1}-\mathbf{w}_s^{t,e}\rangle] \nonumber\\
    &+\frac{\gamma_1}2\mathbb{E}\left[\left\|\mathbf{w}_s^{t,e+1}-\mathbf{w}_s^{t,e}\right\|^2\right].
\end{align}
For $\mathbb{E}[\langle\nabla_{\mathbf{w}_s}F(\mathbf{w}_s^{t,e}),\mathbf{w}_s^{t,e+1}-\mathbf{w}_s^{t,e}\rangle]$, we have
\begin{align}
    &\mathbb{E}[\langle\nabla_{\mathbf{w}_s}F(\mathbf{w}_s^{t,e}),\mathbf{w}_s^{t,e+1}-\mathbf{w}_s^{t,e}\rangle] \nonumber\\
    &=-\eta\mathbb{E}\big[\big\langle\nabla_{\mathbf{w}_{s}}F\big(\mathbf{w}_{s}^{t,e}\big),\nabla_{\mathbf{w}_{s}}f_{s}\big(\mathbf{w}_{s}^{t,e}\big)\big\rangle\big] \nonumber\\
    &=-\eta\mathbb{E}\big[\big\langle\nabla_{\mathbf{w}_{s}}F\big(\mathbf{w}_{s}^{t,e}\big),\nabla_{\mathbf{w}_{s}}F_{s}\big(\mathbf{w}_{s}^{t,e}\big)\big\rangle\big] \nonumber\\
    &=\frac{\eta}{2}\mathbb{E}\left[\|\nabla_{\mathbf{w}_{s}}F(\mathbf{w}_{s}^{t,e})-\nabla_{\mathbf{w}_{s}}F_{s}(\mathbf{w}_{s}^{t,e})\|^{2}\right] \nonumber \\
    &~~-\frac{\eta}{2}\|\nabla_{\mathbf{w}_{s}}F(\mathbf{w}_{s}^{t,e})\|^{2}-\frac{\eta}{2}\mathbb{E}\left[\|\nabla_{\mathbf{w}_{s}}F_{s}(\mathbf{w}_{s}^{t,e})\|^{2}\right] \nonumber \\
    &\leq \frac\eta2\kappa_1^2-\frac\eta2\left\|\nabla_{\mathbf{w}_s}F(\mathbf{w}_s^{t,e})\right\|^2-\frac\eta2\mathbb{E}\left[\left\|\nabla_{\mathbf{w}_s}F_s(\mathbf{w}_s^{t,e})\right\|^2\right].
\end{align}
For $\frac{\gamma_1}2\mathbb{E}\left[\left\|\mathbf{w}_s^{t,e+1}-\mathbf{w}_s^{t,e}\right\|^2\right]$, we have
\begin{align}
    &\frac{\gamma_1}2\mathbb{E}\left[\left\|\mathbf{w}_s^{t,e+1}-\mathbf{w}_s^{t,e}\right\|^2\right]=\frac{\gamma_1\eta^2}2\mathbb{E}\left[\left\|\nabla_{\mathbf{w}_s}f_s(\mathbf{w}_s^{t,e})\right\|^2\right] \nonumber\\
&\leq\frac{\gamma_1\eta^2\sigma^2}{2BQ_s+2\widehat{B}}+\frac{\gamma_1\eta^2}2\mathbb{E}\left[\left\|\nabla_{\mathbf{w}_s}F_s(\mathbf{w}_s^{t,e})\right\|^2\right].
\end{align}
Thus
\begin{align}
    &\mathbb{E}[F(\mathbf{w}_s^{t,e+1})]-F(\mathbf{w}_s^{t,e}) \nonumber\\
    &\leq-\frac{\eta}{2}\left\|\nabla_{{\mathbf{w}_{s}}}F(\mathbf{w}_{s}^{t,e})\right\|^{2}-\frac{\eta}{2}\mathbb{E}\left[\left\|\nabla_{{\mathbf{w}_{s}}}F_{s}(\mathbf{w}_{s}^{t,e})\right\|^{2}\right]\nonumber\\
&+\frac{\gamma_1\eta^2}2\mathbb{E}\left[\left\|\nabla_{\mathbf{w}_s}F_s(\mathbf{w}_s^{t,e})\right\|^2\right]+\frac{\gamma_1\eta^2\sigma^2}{2BQ_s+2\widehat{B}}+\frac\eta2\kappa_1^2.
\end{align}
When $\eta \leq \frac{1}{\gamma_1}$, we have
\begin{align}
    &\mathbb{E}[F(\mathbf{w}_s^{t,e+1})]-F(\mathbf{w}_s^{t,e}) \nonumber\\
    &\leq-\frac\eta2\left\|\nabla_{\mathbf{w}_s}F\left(\mathbf{w}_s^{t,e}\right)\right\|^2+\frac{\gamma_1\eta^2\sigma^2}{2BQ_s+2\widehat{B}}+\frac\eta2\kappa_1^2.
\end{align}
Taking the total expectation and averaging over all rounds, we can obtain the convergence rate:
\begin{align}
\frac{1}{ET} &\sum_{t=0}^{T-1} \sum_{e=0}^{E-1} \mathbb{E} \left[ \|\nabla_{\mathbf{w}_s} F(\mathbf{w}_s^{t,e})\|^2 \right] \nonumber\\
&\leq \mathcal{O} \left( \frac{F(\mathbf{w}_s^0) - F^*}{ET \eta } + \frac{\eta \sigma^2}{BQ_s + \widehat{B}} + \kappa^2_1 \right),
\end{align}
which completes the proof.
\subsubsection{Convergence rate of client-side model:} When Assumption 1 holds, the decrease of the loss function can be bounded as 
\begin{align}\label{lsmooth}
    &\mathbb{E}[F(\mathbf{w}_c^{t+1})]-F(\mathbf{w}_c^t)\nonumber\\
    &\leq\mathbb{E}[\langle\nabla_{\mathbf{w}_c}F(\mathbf{w}_c^t),\mathbf{w}_c^{t+1}-\mathbf{w}_c^t\rangle]+\frac{\gamma_2}2\mathbb{E}[\|\mathbf{w}_c^{t+1}-\mathbf{w}_c^t\|^2].
\end{align}
For $\mathbb{E}[\langle\nabla_{\mathbf{w}_c}F(\mathbf{w}_c^t),\mathbf{w}_c^{t+1}-\mathbf{w}_c^t\rangle]$, we have
\begin{align}\label{innerpro}
    &\mathbb{E}[\langle\nabla_{\mathbf{w}_c}F(\mathbf{w}_c^t),\mathbf{w}_c^{t+1}-\mathbf{w}_c^t\rangle] \nonumber\\
    &=-\eta\sum_{e=0}^{E-1}\mathbb{E}\left[\left\langle\nabla_{\mathbf{w}_{c}}F(\mathbf{w}_{c}^{t}),\sum_{k=1}^{K}\frac1K\nabla_{\mathbf{w}_{c}}F_{k}(\mathbf{w}_{c,k}^{t-\tau^{t},e})\right\rangle\right] \nonumber\\
    &=\frac12\eta\sum_{e=0}^{E-1}\mathbb{E}\left[\left\|\nabla_{\mathbf{w}_c}F(\mathbf{w}_c^t)-\sum_{k=1}^K\frac1K\nabla_{\mathbf{w}_c}F_k(\mathbf{w}_{c,k}^{t-\tau^t,e})\right\|^2\right] \nonumber\\
    &~~~~-\frac12\eta E\left\|\nabla_{\mathbf{w}_c}F(\mathbf{w}_c^t)\right\|^2 \nonumber \\
    &~~~~-\frac12\eta\sum_{e=0}^{E-1}\mathbb{E}\left[\left\|\sum_{k=1}^K\frac1K\nabla_{\mathbf{w}_c}F_k(\mathbf{w}_{c,k}^{t-\tau^t,e})\right\|^2\right].
\end{align}
Define $p_k$ as the probability that client $k$ participates in the training, we have
\begin{align}
    &\mathbb{E}\left[\left\|\nabla_{\mathbf{w}_c}F(\mathbf{w}_c^t)-\sum_{k=1}^K\frac1K\nabla_{\mathbf{w}_c}F_k(\mathbf{w}_{c,k}^{t-\tau^t,e})\right\|^2\right] \nonumber\\
&=\mathbb{E}\left[\left\|\sum_{k=1}^K\frac1K\left(\nabla_{\mathbf{w}_c}F_k\left(\mathbf{w}_{c,k}^{t-\tau^t,e}\right)-\nabla_{\mathbf{w}_c}F_k(\mathbf{w}_c^t)\right)\right\|^2\right] \nonumber\\
&\leq \gamma_2^2\sum_{k=1}^K\frac1K\mathbb{E}\left[\left\|\mathbf{w}_c^t-\mathbf{w}_{c,k}^{t-\tau^t,e}\right\|^2\right] \nonumber\\
& \leq2\gamma_2^2\sum_{k=1}^K\frac{p_k}C\mathbb{E}\left[\left\|\mathbf{w}_c^{t-\tau^t}-\mathbf{w}_{c,k}^{t-\tau^t,e}\right\|^2\right] \nonumber\\
&\quad\quad\quad\quad+2\gamma_2^2\mathbb{E}\left[\left\|\mathbf{w}_c^t-\mathbf{w}_c^{t-\tau^t}\right\|^2\right] \nonumber\\
& \leq2\gamma_2^2\sum_{k=1}^K\frac{p_k}C\mathbb{E}\left[\left\|\mathbf{w}_c^{t-\tau^t}-\mathbf{w}_{c,k}^{t-\tau^t,e}\right\|^2\right] \nonumber\\
&\quad\quad\quad\quad+2\gamma_2^2\tau_{\max}\sum_{j=t-\tau_{\max}}^{t-1}\mathbb{E}\left[\left\|\mathbf{w}_c^{j+1}-\mathbf{w}_c^j\right\|^2\right].
\end{align}
According to \cite{yang2021achieving,wang2023lightweight}, when $\eta\leq\frac1{\sqrt{30}\gamma_2E}$, we have
\begin{align}
    &\sum_{e=0}^{E-1}\mathbb{E}\left[\left\|\mathbf{w}_c^t-\mathbf{w}_{c,k}^{t,e}\right\|^2\right]\nonumber\\
&\leq5E^2\eta^2(\frac{\sigma^2}{B}+6E\kappa_2^2)+30E^3\eta^2\left\|\nabla_{\mathbf{w}_c}F(\mathbf{w}_c^t)\right\|^2.
\end{align}
Thus the first term of \eqref{innerpro} can be bounded as
\begin{align}\label{first}
    &30\gamma_2^2E^3\eta^3\mathbb{E}\left[\left\|\nabla_{\mathbf{w}_c}F(\mathbf{w}_c^{t-\tau^t})\right\|^2\right] \nonumber\\
    &\quad\quad+\gamma_2^2\eta E\tau_{\max}\sum_{j=t-\tau_{\max}}^{t-1}\mathbb{E}\left[\left\|\mathbf{w}_c^{j+1}-\mathbf{w}_c^j\right\|^2\right] \nonumber\\
    &\quad\quad+\frac{5\gamma_2^2E^2\eta^3\sigma^2}{B}+30\gamma_2^2E^3\eta^3\kappa_2^2.
\end{align}
For the last term of \eqref{lsmooth}, we have
\begin{align}\label{wcj}
    &\frac{\gamma_2}2\mathbb{E}\left[\left\|\mathbf{w}_c^{j+1}-\mathbf{w}_c^j\right\|^2\right]
    \nonumber \\
    &\leq\gamma_2\eta^2E\sum_{e=1}^{E-1}\mathbb{E}\left[\left\|\frac1C\sum_{k\in\mathcal{C}^j}\nabla_{\mathbf{w}_c}F_k\left(\mathbf{w}_{c,k}^{j-\tau^j,e}\right)\right.\right. \nonumber\\
    &\quad\quad\quad-\left.
    \left.\sum_{k=1}^K\frac1K\nabla_{\mathbf{w}_c}F_k\left(\mathbf{w}_{c,k}^{j-\tau^j,e}\right)\right\|^2\right] \nonumber\\
    &\quad\quad\quad+\gamma_2\eta^2E\sum_{e=1}^{E-1}\mathbb{E}\left[\left\|\sum_{k=1}^K\frac1K\nabla_{\mathbf{w}_c}F_k\left(\mathbf{w}_{c,k}^{j-\tau^j,e}\right)\right\|^2\right] \nonumber\\
    &\quad\quad\quad+\frac{\gamma_2\eta^2E^2\sigma^2}{2B}.
\end{align}
Then the first term of \eqref{wcj} can be bounded as
\begin{align}
    &\gamma_2\eta^2E\sum_{e=1}^{E-1}\mathbb{E}\left[\left\|\frac1C\sum_{k\in\mathcal{C}}\nabla_{\mathbf{w}_c}F_k\left(\mathbf{w}_{c,k}^{j-\tau^j,e}\right)\right.\right. \nonumber\\
    &\quad\quad\quad\quad-\left.
    \left.\sum_{k=1}^K\frac1K\nabla_{\mathbf{w}_c}F_k\left(\mathbf{w}_{c,k}^{j-\tau^j,e}\right)\right\|^2\right] \nonumber\\
    &\leq\gamma_2\eta^2E\sum_{k\in\mathcal{C}^j}\frac1C\sum_{e=1}^{E-1}\mathbb{E}\Bigg\{\mathbb{E}\Bigg(\left\|\nabla_{\mathbf{w}_c}F_k\left(\mathbf{w}_{c,k}^{j-\tau^j,e}\right.\right) \nonumber\\
    &\quad\quad\quad\quad-\left.\sum_{r=1}^K\frac1K\nabla_{\mathbf{w}_c}F_r\left(\mathbf{w}_{c,r}^{j-\tau^j,e}\right)\right\|^2|\mathcal{C}\Bigg)\Bigg\} \nonumber\\
    &\leq\gamma_2\eta^2E\sum_{k\in\mathcal{C}^j}\frac1C\sum_{e=1}^{E-1}\mathbb{E}\Bigg\{\mathbb{E}\Bigg(\left\|\nabla_{\mathbf{w}_c}F_k\left(\mathbf{w}_{c,k}^{j-\tau^j,e}\right.\right) \nonumber\\
    &\quad-\nabla_{\mathbf{w}_c}F\left(\mathbf{w}_c^{j-\tau^j}\right)\Bigg\|^2|\mathcal{C}\Bigg)\Bigg\}+3\gamma_2\eta^2E^2\kappa_2^2\nonumber\\
    &\quad+3\gamma_2\eta^2E\sum_{e=1}^{E-1}\mathbb{E}\left[\left\|\sum_{k=1}^K\frac1K\Bigg(\nabla_{\mathbf{w}_c}F_k\left(\mathbf{w}_c^{j-\tau^j}\right)\right.\right.\nonumber\\
    &\quad-\nabla_{\mathbf{w}_c}F_k\left(\mathbf{w}_{c,k}^{j-\tau^j,e}\right)\Bigg)\Bigg\|^2\Bigg]\nonumber\\
    &\leq180\gamma_2^3\eta^4E^4\mathbb{E}\left[\left\|\nabla_{\mathbf{w}_c}F\left(\mathbf{w}_c^{j-\tau^j}\right)\right\|^2\right]\nonumber\\
    &\quad+\frac{30\gamma_2^3\eta^4E^3\sigma^2}{B}+180\gamma_2^3\eta^4E^4\kappa_2^2+3\gamma_2\eta^2E^2\kappa_2^2.
\end{align}
Let 
\begin{align}
&P=\frac{\eta^2E^2\sigma^2+60\gamma_2^2\eta^4E^3\sigma^2}{B}\nonumber\\
&\quad\quad\quad\quad+360\gamma_2^2\eta^4E^4\kappa_2^2+6\eta^2E^2\kappa_2^2\nonumber\\
&R(t-\tau^t)=\sum_{e=1}^{E-1}\mathbb{E}\left[\left\|\sum_{k=1}^K\frac1K\nabla_{\mathbf{w}_c}F_k\left(\mathbf{w}_{c,k}^{t-\tau^t,e}\right)\right\|^2\right],\nonumber
\end{align}
then the last term of \eqref{lsmooth} can be bounded as 
\begin{align}\label{second}
    &\frac{\gamma_2}2\mathbb{E}\left[\left\|\mathbf{w}_c^{j+1}-\mathbf{w}_c^j\right\|^2\right]
    \nonumber \\
    &\leq180\gamma_2^3\eta^4E^4\mathbb{E}\left[\left\|\nabla_{\mathbf{w}_c}F\left(\mathbf{w}_c^{j-\tau^j}\right)\right\|^2\right]\nonumber \\
    &\quad+\gamma_2\eta^2ER(t-\tau^t)+0.5\gamma_2 P.
\end{align}
Then for $\gamma_2^2\eta E\tau_{\max}\sum_{j=t-\tau_{\max}}^{t-1}\mathbb{E}\left[\left\|\mathbf{w}_c^{j+1}-\mathbf{w}_c^j\right\|^2\right]$, we have
\begin{align}\label{third}
    &\gamma_2^2\eta E\tau_{\max}\sum_{j=t-\tau_{\max}}^{t-1}\mathbb{E}\left[\left\|\mathbf{w}_c^{j+1}-\mathbf{w}_c^j\right\|^2\right] \nonumber\\
    &\leq360\gamma_2^4\eta^5E^5\tau_{\max}\sum_{j=t-\tau_{\max}}^{t-1}\mathbb{E}\left[\left\|\nabla_{\mathbf{w}_c}F\left(\mathbf{w}_c^{j-\tau^j}\right)\right\|^2\right]\nonumber\\
    &+2\gamma_2^2\eta^3E^2\tau_{\max}\sum_{j=t-\tau_{\max}}^{t-1}R(j-\tau^j)+\gamma_2^2\eta E\tau_{\max}^2P.
\end{align}
Incorporating \eqref{first}, \eqref{second}, and \eqref{third} into \eqref{lsmooth}, and then taking the total expectation and averaging over all rounds, we can obtain the convergence rate:
\begin{align}\label{integrate}
    &\frac{\mathbb{E}[F(\mathbf{w}_c^{T-1})]-F(\mathbf{w}_c^0)}T \nonumber\\
    &\leq \Big(-\frac12\eta E+30\gamma_2^2E^3\eta^3\tau_{\max}+180\gamma_2^3\eta^4E^4\tau_{\max}\nonumber\\
    &+360\gamma_2^4\eta^5E^5\tau_{\max}^3\Big)\frac1T\sum_{t=0}^{T-1}\mathbb{E}\left[\left\|\nabla_{\mathbf{w}_c}F(\mathbf{w}_c^t)\right\|^2\right]\nonumber\\
    &+\frac{5\gamma_2^2E^2\eta^3\sigma^2}{B}+30\gamma_2^2E^3\eta^3\kappa_2^2+0.5\gamma_2P+\gamma_2^2\eta E\tau_{\max}^2P\nonumber\\
    &+\frac1T\sum_{t=0}^{T-1}(2\gamma_2^2\eta^3E^2\tau_{\max}^3+\gamma_2\eta^2E\tau_{\max}-0.5\eta\tau_{\max})R(t).
\end{align}
By assuming that $\eta\leq\frac1{20\gamma E\sqrt{\tau_{\max}}}$, \eqref{integrate} can be bounded as:
\begin{align}\label{integrate2}
   &\frac{\mathbb{E}[F(\mathbf{w}_c^{T-1})]-F(\mathbf{w}_c^0)}T\nonumber\\
    &\leq\left(-\frac14\eta E\right)\frac1T\sum_{t=0}^{T-1}\mathbb{E}\left[\left\|\nabla_{\mathbf{w}_c}F(\mathbf{w}_c^t)\right\|^2\right]+\frac{5\gamma_2^2E^2\eta^3\sigma^2}{B}\nonumber\\
    &\quad+30\gamma_2^2E^3\eta^3\kappa_2^2+0.5\gamma_2P+\gamma_2^2\eta E\tau_{\max}^2P.
\end{align}
Neglecting the higher-order terms of $\eta$, \eqref{integrate2} can be simplified as
\begin{align}
    &\frac{1}{T} \sum_{t=0}^{T-1}  \mathbb{E} \left[ \|\nabla_{\mathbf{w}_c} F(\mathbf{w}_c^t)\|^2 \right] 
\leq \mathcal{O} \Bigg( \frac{F(\mathbf{w}_s^0) - F^*}{ET \eta } \nonumber \\
&~~~~~~~~+ \left( \frac{\sigma^2}{B} + \kappa_2^2 \right) \eta E + \left( \frac{\sigma^2}{B} + \kappa_2^2 \right) \eta^2 E^2 \tau_{\max}^2 \Bigg),
\end{align}
which completes the proof.

\subsection{E. More Experimental Details}
We conducted our experiments using two RTX 4080 GPUs. For all datasets, we used the SGD optimizer with a learning rate of $0.01$, momentum of $0.9$, and weight decay of $0.0005$. For each training task, we ran experiments with random seeds of $2023$, $1998$, and $1125$. Full details are provided in the code.

\subsection{F. Ablation Study on Client Number}
\begin{table}[t]
\resizebox{\linewidth}{!}{
\begin{tabular}{cccccc}
\toprule
\multirow{2}{*}{}        & \multirow{2}{*}{Method} & \multicolumn{2}{c}{CIFAR-10}             & \multicolumn{2}{c}{CINIC-10}             \\ 
\cmidrule(lr){3-4}\cmidrule(lr){5-6}
                         &                         & $\text{s}=2$   & $\alpha=0.1$       & $\text{s}=2$   & $\alpha=0.1$       \\ \midrule
\multirow{4}{*}{$K=50$}  & FedAvg                  & $71.38$ & $74.61$ & $54.11$ & $49.75$ \\
                         & FedBuff                 & $70.91$ & $76.45$ & $49.26$ & $52.68$ \\
                         & CA$^2$FL                & $80.78$ & $78.53$ & $65.97$ & $63.55$ \\ \cmidrule(lr){2-6} 
                         & Ours                    & $\mathbf{82.53}$ & $\mathbf{81.42}$ & $\mathbf{67.79}$ & $\mathbf{65.82}$ \\ \midrule
\multirow{4}{*}{$K=100$} & FedAvg                  & $63.66$ & $76.44$ & $51.91$ & $47.68$ \\
                         & FedBuff                 & $67.63$ & $73.69$ & $49.56$ & $51.68$ \\
                         & CA$^2$FL                & $80.25$ & $78.22$ & $64.53$ & $64.21$ \\ \cmidrule(lr){2-6}  
                         & Ours                    & $\mathbf{82.56}$ & $\mathbf{81.08}$ & $\mathbf{66.74}$ & $\mathbf{64.67}$ \\ \bottomrule
\end{tabular}}
\caption{Test accuracy (\%) under different number of clients.}
\label{table3}
\end{table}
In this section, we evaluate the scalability of the proposed GAS by conducting experiments with an increased number of clients. Specifically, we consider scenarios with $50$ clients ($K=50$) and $100$ clients ($K=100$). The number of participating clients in each global iterations is set to $10$, with local iterations and global iterations set to $10$ and $2000$, respectively. The results are presented in Table \ref{table3}, which indicates that GAS not only scales well with an increasing number of clients but also consistently outperforms the baseline methods in terms of model accuracy. This demonstrates the robustness and efficiency of GAS in large-scale FL environments characterized by significant data heterogeneity.

\subsection{G. Ablation Study on Splitting layer}
\begin{table}[t]
\centering
\resizebox{\linewidth}{!}{
\begin{tabular}{ccccc}
\toprule
Layer    & $L_1$   & $L_2$   & $L_3$   & $L_4$ \\ \midrule
AlexNet & $82.71_{\pm 1.73}$ & $82.50_{\pm 0.97}$ & $82.07_{\pm 0.65}$ & $72.02_{\pm 4.39}$ \\
VGG16 & $87.63_{\pm 0.84}$ & $84.27_{\pm 0.82}$ & $76.36_{\pm 3.69}$ & $79.01_{\pm 2.92}$ \\ \bottomrule
\end{tabular}}
\caption{Test accuracy (\%) under different split layers.}
\label{table4}
\end{table}

In this section, we evaluate the impact of different split layers on the performance of GAS on CIFAR-10 under extreme data heterogeneity conditions with $\text{shard}= 3$. Specifically, we consider the performance of AlexNet \cite{krizhevsky2012imagenet} with splits at the $3$-rd, $6$-th, $7$-th, and $11$-th layers, and VGG16 \cite{simonyan2014very} with splits at the $10$-th, $17$-th, $24$-th, and $31$-st layers, where the split layers from shallow to deep is denoted as $L_1$ to $L_4$. The global iterations are set to $1000$ and the experimental results are shown in Table \ref{table4}. The experimental results indicate that the model achieves the highest accuracy when the split layer is the shallowest. This finding demonstrates that deploying more layers of the model on the server for centralized updates can enhance performance under extreme data heterogeneity. However, shallower split layers often result in higher activation dimensions, making the estimation of activation distributions more challenging and potentially impacting model performance \cite{tang2024fedimpro}. This is evident in the VGG16 model, where the output activation dimension at the shallower $L_3$ split layer is $2 \times 2 \times 512$, resulting in lower accuracy compared to the $L_4$ split layer, which has an output activation dimension of $1 \times 1 \times 512$. Therefore, the split layer should be carefully chosen. On one hand, selecting shallower split layers can leverage the performance benefits of centralized updates. On the other hand, it is essential to avoid split layers with large output dimensions to mitigate the complexity of activation distribution estimation and its adverse effects on model performance.

\begin{table}[t]
\centering
\begin{tabular}{cccc}
\toprule
\multirow{2}{*}{}          & \multirow{2}{*}{Method} & \multicolumn{2}{c}{CIFAR-10}                              \\ \cmidrule(lr){3-4}
                           &                         & $\text{s}=2$                & $\alpha=0.1$                \\ \midrule
\multirow{4}{*}{ViT-small} & FedAvg                  & $70.85_{\pm 0.68}$          & $68.95_{\pm 1.36}$          \\
                           & FedBuff                 & $58.46_{\pm 2.18}$          & $59.84_{\pm 3.64}$          \\
                           & CA$^2$FL                & $68.27_{\pm 0.11}$          & $70.58_{\pm 1.20}$          \\ \cmidrule(lr){2-4} 
                           & Ours                    & $\mathbf{79.81}_{\pm 1.43}$ & $\mathbf{72.89}_{\pm 4.75}$ \\ \midrule
\multirow{4}{*}{ResNet18}  & FedAvg                  & $34.47_{\pm 2.51}$          & $75.39_{\pm 7.04}$          \\
                           & FedBuff                 & $31.83_{\pm 3.45}$          & $72.73_{\pm 6.71}$          \\
                           & CA$^2$FL                & $37.81_{\pm 8.77}$          & $76.28_{\pm 7.16}$          \\ \cmidrule(lr){2-4} 
                           & Ours                    & $\mathbf{83.58}_{\pm 5.62}$ & $\mathbf{80.73}_{\pm 1.21}$ \\ \bottomrule
\end{tabular}
\caption{Test accuracy (\%) under different model structures.}
\label{table5}
\end{table}

\subsection{H. Ablation Study on Model Structures}
We conduct supplementary experiments using ViT-small \cite{han2022survey} and ResNet18 \cite{he2016deep}. In the ViT-small experiments, we employ a compact Vision Transformer architecture consisting of $6$ Transformer layers, each with $8$ attention heads and a hidden MLP dimension of $512$. We set the dropout rate to $0.1$ for both the Transformer layers and the embedding layer. The split point of the model is set after the first Transformer layer, with a learning rate of $0.0001$, optimized using the Adam optimizer. In the ResNet18 supplementary experiments, the model is split after the first residual layer, and we use a learning rate of $0.005$ with the SGD optimizer. The experimental results, as shown in the Table \ref{table5}, demonstrate that the proposed GAS exhibits good scalability and outperforms benchmarks across various levels of data heterogeneity.

\subsection{I. Enhancing Privacy in GAS}
In all SL and SFL methods, including the proposed GAS, there exists a risk of privacy leakage due to the exchange of intermediate activations between clients and the server. Specifically, these activations may contain sensitive information about the original data and an attacker can leverage the Model Inversion (MI) attack to repeatedly access and analyze the activations, gradually reconstructing the original data during the training phase. To address this issue, our proposed GAS can integrate with existing privacy-preserving mechanisms of SFL, such as NoPeek \cite{vepakomma2020nopeek} and ResSFL \cite{li2022ressfl}, which implement attacker-aware training to counteract MI attacks.
Specifically, we introduce an inversion score regularization term as
\begin{align}
f_k(\mathbf{w}_c)=&l(\mathbf{w}_s;h(\mathbf{w}_c;\tilde{\mathcal{D}}_k))\nonumber\\
&
+\gamma \mathbb{R}(\mathcal{L}(\mathbf{w}_v;h(\mathbf{w}_c;\tilde{\mathcal{D}}_k)),\tilde{\mathcal{D}}_k),
\end{align}
where $\mathbf{w}_v$ is the simulated inversion model, and $\mathbb{R}$ is the score function used to evaluate the quality of the reconstructed images compared to the ground-truth images $\mathcal{D}_k$. This term increases the difficulty for an attacker to reconstruct the original data from the activations, thereby reducing the risk of sensitive data leakage.
\bibliography{aaai25}
\end{document}